\documentclass[10pt,twocolumn,letterpaper]{article}

\usepackage{wacv}              

\usepackage[accsupp]{axessibility} 
\usepackage{graphicx}
\usepackage{amsmath}
\usepackage{amssymb}
\usepackage{booktabs}

\usepackage{multirow}
\usepackage{times}
\usepackage{epsfig}
\usepackage{amsfonts}
\usepackage{comment}
\usepackage{color}
\usepackage[dvipsnames]{xcolor}
\usepackage{balance}
\usepackage{enumitem}
\usepackage{adjustbox}
\usepackage{color, colortbl}

\definecolor{LightCyan}{rgb}{0.88,0.88,1}
\definecolor{arylideyellow}{rgb}{0.91, 0.84, 0.42}
\definecolor{blond}{rgb}{0.98, 0.94, 0.75}
\definecolor{Gray}{gray}{0.9}
\definecolor{babypink}{rgb}{0.96, 0.76, 0.76}
\definecolor{lightcyan}{rgb}{0.88, 1.0, 1.0}
\definecolor{piggypink}{rgb}{0.99, 0.87, 0.9}
\definecolor{coolblack}{rgb}{0.0, 0.18, 0.39}
\definecolor{darkmidnightblue}{rgb}{0.0, 0.2, 0.4}

\usepackage{hhline}
\usepackage{makecell}

\usepackage{algorithm}
\usepackage{algpseudocode}
\usepackage{wrapfig,lipsum,booktabs}
\usepackage{import}
\usepackage{caption}

\usepackage[pagebackref,breaklinks,colorlinks]{hyperref}

\usepackage[capitalize]{cleveref}
\crefname{section}{Sec.}{Secs.}
\Crefname{section}{Section}{Sections}
\Crefname{table}{Table}{Tables}
\crefname{table}{Tab.}{Tabs.}

\clearpage 
\appendix

\def\wacvPaperID{2142} 
\def\confName{WACV}
\def\confYear{2024}

\begin{document}

\title{Open-Set Object Detection By Aligning Known Class Representations}
\author{ Hiran Sarkar$^1$ $\quad$ Vishal Chudasama$^1$ $\quad$ Naoyuki Onoe$^1$ \\ Pankaj Wasnik$^1$\thanks{Corresponding author} $\quad$ Vineeth N Balasubramanian$^2$ \\
$^{1}$Sony Research India$\quad$  $^{2}$Indian Institute of Technology Hyderabad\\
{\tt\small \{hiran.sarkar,vishal.chudasama1,naoyuki.onoe,pankaj.wasnik\}@sony.com, vineethnb@cse.iith.ac.in}
}
\maketitle
\begin{abstract}
Open-Set Object Detection (OSOD) has emerged as a contemporary research direction to address the detection of unknown objects. Recently, few works have achieved remarkable performance in the OSOD task by employing contrastive clustering to separate unknown classes. In contrast, we propose a new semantic clustering-based approach to facilitate a meaningful alignment of clusters in semantic space and introduce a class decorrelation module to enhance inter-cluster separation. Our approach further incorporates an object focus module to predict objectness scores, which enhances the detection of unknown objects. Further, we employ i) an evaluation technique that penalizes low-confidence outputs to mitigate the risk of misclassification of the unknown objects and ii) a new metric called \textit{HMP} that combines known and unknown precision using harmonic mean. Our extensive experiments demonstrate that the proposed model achieves significant improvement on the MS-COCO \& PASCAL VOC dataset for the OSOD task. 
\end{abstract}

\begin{figure}
    \centering
        \subfloat[Illustration of cluster evolution through incremental component inclusion. Colored dots denote known classes, black dots indicate unknown classes. With semantic clustering, known class clusters align semantically. Class decorrelation further enhances cluster separation.]{\includegraphics[width=0.99\columnwidth, height = 0.27\textheight]{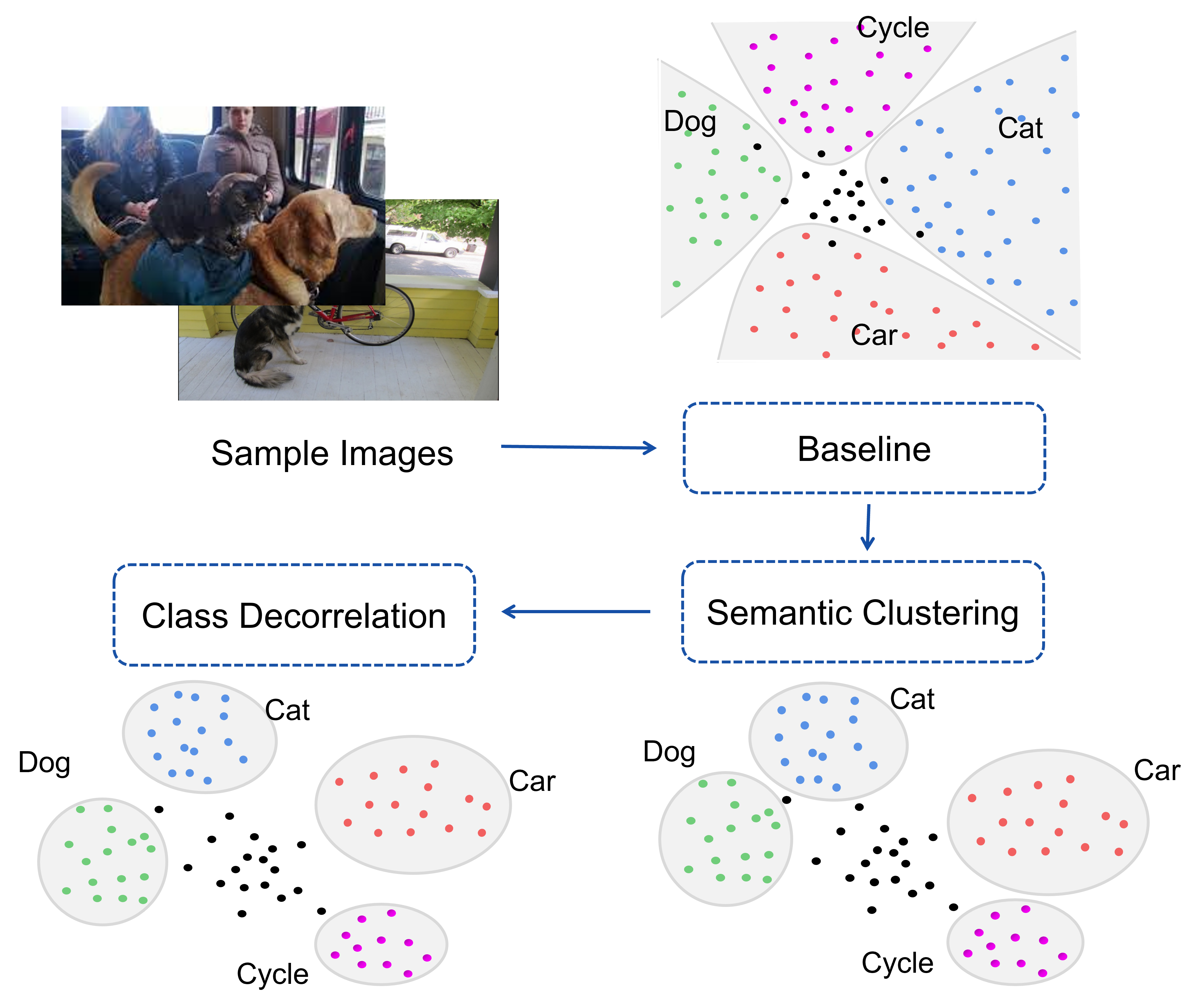}} \\
        \subfloat[Visual comparison between the proposed and OpenDet \cite{OpenDet} methods.]{\includegraphics[width=0.99\columnwidth,height = 0.24\textheight]{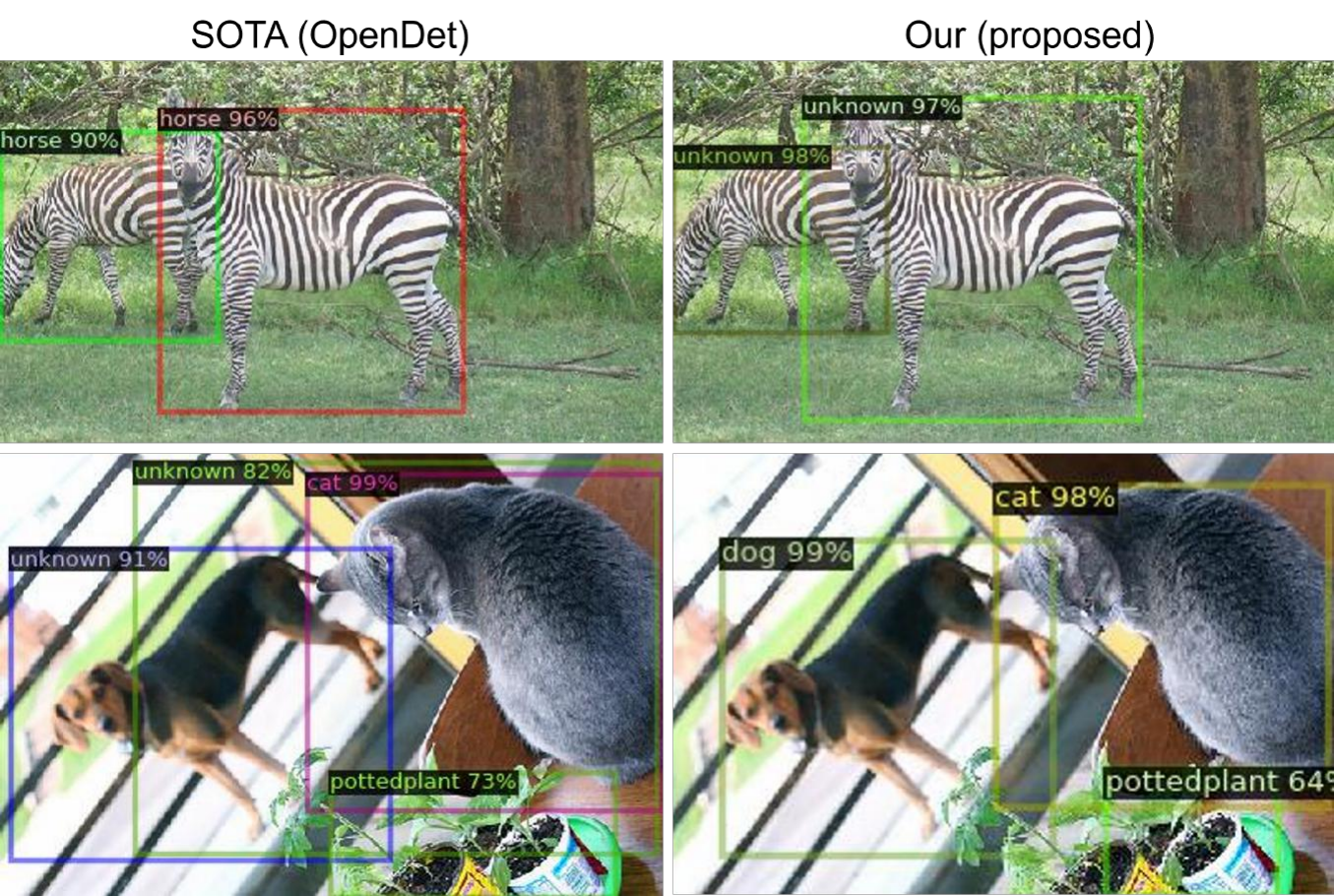}} 
        \vspace{-0.7em}
\caption{Effectiveness of semantic clustering, class decorrelation, and visual comparison with OpenDet \cite{OpenDet}.} \label{fig:intro}
\end{figure}
\section{Introduction}
\label{sec:intro}
Object detection task has seen significant advancements in the past decade. However, many of the current object detectors often fail to localize or classify objects of novel or unseen classes. To address this, Open-Set Object Detection (OSOD) has been introduced, which aims to detect new or unidentified objects as ``unknown" class along with the known objects with their respective categories. One of the key challenges in OSOD is the issue of misclassification of unknown class with high confidence. This is especially for the case of unknown objects that exhibit semantic closeness to a known class. For instance, an open-set detector that is trained on VOC classes \cite{PascalVOC} might misclassify an unknown class, e.g., ``zebra" to a close known class like ``horse". Such a misclassification has been observed in contrastive clustering-based previous OSOD works \cite{ORE,OpenDet,OpenSetRCNN} due to the semantic proximity of known and unknown classes.
Moreover, the methods utilizing classification-based Region Proposal Network (RPN) to predict objectness, obtain lower performance on objects with different geometric attributes than the known classes. 

This study proposes a new framework to address the aforementioned concerns. To tackle the issue of unknown misclassification, we introduce a new semantic clustering module that aligns region proposal features with their respective semantic class embeddings. This enables the detector to establish meaningful class decision boundaries, thereby preventing unknown misclassification, as illustrated in Figure~\ref{fig:intro}(a). Moreover, we impose an orthogonality constraint on the features to ensure a clear separation of the clusters. For this purpose, we introduce a new class decorrelation module inspired by \cite{IDFD} that utilizes the feature decorrelation-based softmax-formulated orthogonality constraint on the cluster features. This module facilitates an increase in inter-class cluster distance, yielding improved unknown separation, as demonstrated in Figure~\ref{fig:intro}(a).

Previous approaches \cite{OpenDet,ORE} employ binary classification-based objectness prediction that tends to overfit on the training categories and constraints the effectiveness of RPN, especially in OSOD setting as discussed in \cite{ctrness, OpenSetRCNN}. 
We alleviate this constraint with object focus loss, that learns objectness with the help of centerness and classification-based objectness loss. Here, the centerness helps the detector to predict how far a proposal is from a ground-truth bounding box. This enables a more robust learning of the RPN, as it prevents overfitting on the training categories by learning from object cues, such as location, geometry and other spatial relationships. This, in turn, facilitates easier and unconstrained detection of unknown objects.
The proposed model accurately identifies unseen objects as unknown class and enhances detection performance compared to previous state-of-the-art (SOTA) methods. In Figure~\ref{fig:intro}(b), results obtained from the proposed method and OpenDet \cite{OpenDet} are illustrated that demonstrates our method detects the unknown classes accurately as compared to OpenDet \cite{OpenDet}. For instance, the proposed model accurately predicts the ``zebra" as an ``unknown" object, while OpenDet \cite{OpenDet} fails and identifies it as a ``horse". Furthermore, in another example, OpenDet \cite{OpenDet} predicts ``dog" as an ``unknown" class, where our model correctly identifies its class. We summarize the contributions of this paper as follows.
\vspace{-0.5em}
\begin{itemize}[leftmargin=*]
\setlength\itemsep{-0.4em}
    \item We propose a new OSOD framework which aligns class representations effectively and detects the unknown objects accurately.
    \item We introduce i) a novel semantic clustering module to group the features in the semantic space that facilitates improved cluster boundary separation, especially between semantically similar objects, ii) a class decorrelation module to further encourage separation of the formed clusters, iii) a new loss known as object focus loss to enable a more resilient learning process of RPN that facilitates the unconstrained detection of unknown objects.
    \item A new evaluation technique, entropy thresholding, is employed to penalize low-confidence outputs, thereby mitigating the risk of misclassifying the unknown objects as known classes. In addition, a new evaluation metric, Harmonic Mean Precision (\textit{HMP}), is employed to combine the precision scores of known and unknown objects.
    \item We performed extensive experiments on benchmark datasets, showing significant improvements over prior work. We also conducted various ablation studies to validate the usefulness of the proposed method.
\end{itemize}
\vspace{-1em}

\begin{figure*}[t!]
    \centering
    \includegraphics[width = 0.96\linewidth,height = 0.42\textheight]{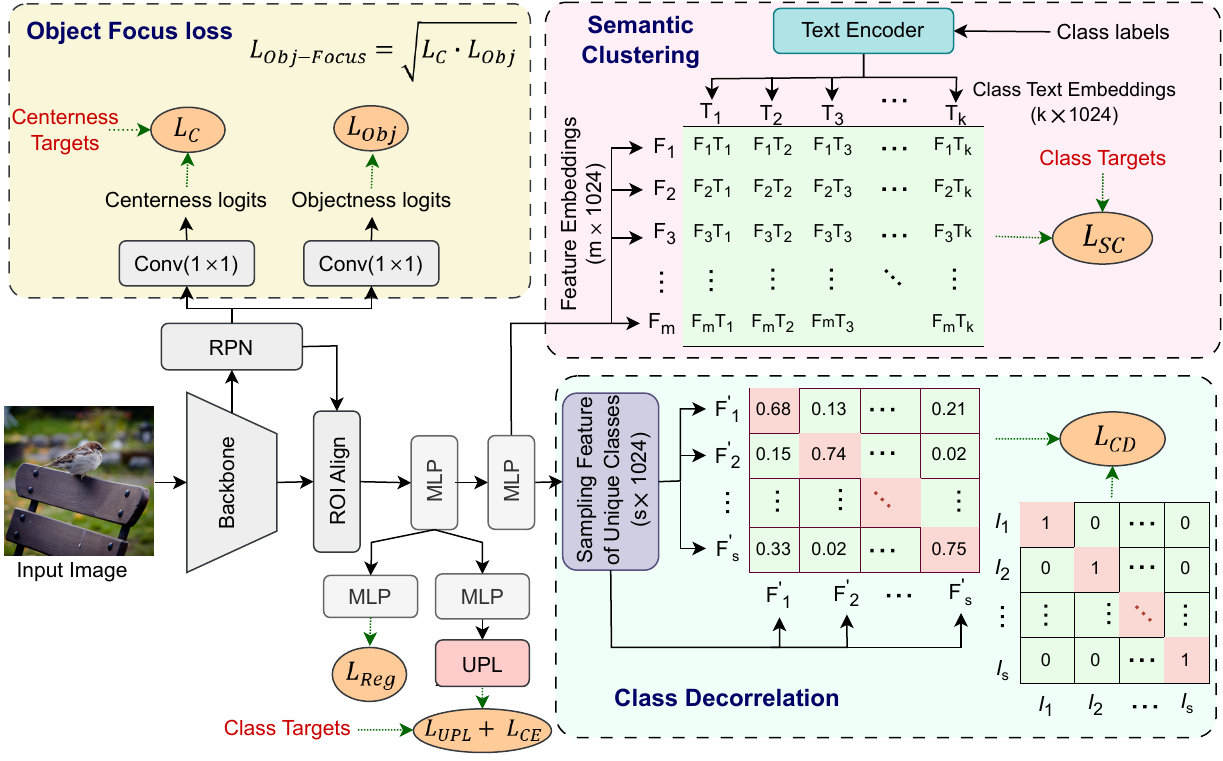}
    \vspace{-1em}
    \caption{\small \textbf{Overview of our proposed method}. \textcolor{darkmidnightblue}{\textbf{Object Focus loss:}} Object focus loss is a combination of $L_C$ (classification free loss) and $L_{obj}$ (classification based loss). \textcolor{darkmidnightblue}{\textbf{Semantic Clustering:}} $\{T_1, T_2, T_3, \ldots, T_k\}$ represents the class embeddings of $k$ classes and $\{F_1, F_2, F_3, \ldots, F_m\}$ represents $m$ feature embeddings. Each $F_i$ gets aligned with its corresponding class embedding. \textcolor{darkmidnightblue}{\textbf{Class Decorrelation:}} We sample one feature per unique class to create $\{F'_1, F'_2, F'_3, \ldots, F'_s\}$, where $s$ represents the total count of unique classes within an iteration. These sampled features are subsequently orthogonalized against the remaining features to ensure orthogonality.}
    \label{fig:method}
    \vspace{-1em}
\end{figure*}
\section{Related Works}
\label{sec:relatedworks}
\textbf{Open-Set Recognition (OSR)} aims to identify the known objects along with unknown or novel objects that were not seen during the training phase. 
Bendale \textit{\textit{et al.}} \cite{OSR5} were the first to introduce a deep learning-based OSR method. Subsequently, several reconstruction-based methods have been proposed to enhance the performance of the OSR task. Few works \cite{OSR6, OSR7} have employed the generative adversarial network to generate potential open-set images to train an open-set classifier. While other approaches \cite{OSR8, OSR9, OSR10} utilized the auto-encoder to recover latent features and identify unknown class by reconstruction errors. 

\textbf{Open-Set Object Detection (OSOD)} is an extension of OSR that aims to detect unseen object as unknown. Dhamija \textit{et al.} \cite{WI} have first formalized OSOD and found that the performance of most detectors is exaggerated in open-set conditions. Joseph \textit{et al.} \cite{ORE} proposed ORE method by introducing an energy-based unknown identifier. Subsequently, several works have been proposed \cite{OWOD1, OWOD2, OW-DETR, UC-OWOD, PROB} to improvise the performance of the ORE model. 
Recently, Gupta \textit{et al.} \cite{OW-DETR} adapted the Deformable DETR model \cite{DETR} for the open world objective and introduced OW-DETR. 
Zohar \textit{et al.} \cite{PROB} proposed a method by integrating the probabilistic objectness into the deformable DETR model \cite{DETR}. Miller \textit{et al.} \cite{DS} have implemented the dropout sampling technique to estimate uncertainty in object detection, aiming to mitigate open-set errors. Recently, Han \textit{et al.} \cite{OpenDet} have introduced contrastive feature learner to encourage compact features of known classes and unknown probability learner to separate known and unknown classes. Zhou \textit{et al.} \cite{OpenSetRCNN} have enhanced the generalization ability for unknown object proposals using a classification-free RPN. These methods \cite{ORE, OpenDet, OpenSetRCNN} adopt a contrastive-based clustering approach to distinguish the unknown objects from the known clusters. However, these methods underperform in cases where an unknown object is semantically closer to a known class. To address this issue, we propose semantic clustering and class decorrelation modules that aims to learn a clear cluster boundary between semantically similar cluster and encourage separation among them. 

\textbf{Classification-free object detection (CFOD)} has emerged recently which focuses on object detection by learning general objectness features rather than relying on class information. This allows to detect previously unseen object classes using these learned features. Recently, Kim \textit{et al.} \cite{ctrness} incorporates centerness features to learn objectness, demonstrating that substituting classification-based loss with a classification-free variant enhances performance for open world proposals. In \cite{OpenSetRCNN}, Zhou \textit{et al.} utilized a classification-free RPN approach for detecting unknown object proposals. Wu \textit{et al.} \cite{ctrness_rel} investigated the direct integration of CFOD into an open-set setting, resulted in a 70\% decrease in unknown object recall.  Inspired by these findings, we propose a new object focus loss that uses a combination of classification-free and classification-based object detection for learning objectness.
\\
\vspace{-0.5em}
\section{Proposed Framework}
\label{sec:methodology}
\vspace{0em}
\subsection{Problem Statement \& Notations}
Given an object detection training dataset $D_{tr} = \{(x, y), x \in X_{tr}, y \in Y_{tr}\}$ with known classes \(C_k = \{c_1, c_2, ..., c_k\}\) and testing dataset $D_{te} = \{(x', y'), x' \in X_{te}, y' \in Y_{te}\}$ containing $k$ known classes ($C_k$) as well as $u$ unknown classes ($C_u$), the objective of OSOD is to accurately detect all known objects belonging to $C_k$, while also identifying novel objects as ``unknown" class. In this context, $X_{tr}$ and $X_{te}$ represent the input images of training and testing dataset, respectively, while $Y_{tr}$ and $Y_{te}$ are set of training and testing labels containing corresponding class labels and bounding boxes. As it is unfeasible to enumerate infinite unknown classes, we denote unknown classes as $C_u = c_{k+1}$. 
The proposed framework introduces three alignment modules to effectively segregate the known class clusters and detect unknown classes accurately. (i) The CLIP-based semantic clustering module that facilitates the formation of clusters in the semantic space. (ii) The class decorrelation module, which enforces an orthogonality constraint among features of different clusters and helps to separate the clusters. (iii) The object focus loss that enhances the unknown detection capabilities. Detailed explanations of these modules are discussed in the subsequent subsections.

\subsection{Semantic Clustering}
Contrastive learning has significantly enhanced the deep clustering of images by capturing distinct visual attributes customized to each instance \cite{CC1, CC2, CC3, CC4, CC5}. Nonetheless, its capacity to explicitly infer class decision boundaries still needs to be explored. This can be attributed to the absence of class sensitivity within the instance discrimination strategy, leading to clusters in the feature space that are not effectively aligned with class decision boundaries. This issue is further compounded when applied in the OSOD setting, where an unknown object is misclassified to its semantically closer known class.


To address this issue, we have introduced a semantic clustering module inspired by CLIP \cite{CLIP}. This module facilitates the formation of clusters in the semantic space, thereby establishing clear semantic decision boundaries. Consequently, the module mitigates the confusion between semantically similar objects, thus minimizing the misclassification of unknown objects to known classes. In this regard, we utilize the CLIP-based text encoder to generate label embeddings $\{T_1, T_2, T_3, \ldots, T_k\}$ that correspond to $k$ ground truth VOC classes \cite{PascalVOC}. Subsequently, a MLP layer is appended after the ROI align module to generate 1024-dimensional image features $\{F_1, F_2, F_3, \ldots, F_m\}$ for $m$ sampled proposals. For each proposal feature $F_i$, we calculate the cosine similarity with each text embedding. The cosine similarity between the $i^{th}$ image feature and the $j^{th}$ label feature is then calculated as
\begin{equation}
    \text{cos-sim}_{ij} = \frac{F_i \cdot T_j}{\|F_i\| \cdot \|T_j\|}, 
\end{equation}
The resulting output (i.e., $\text{cos-sim}_{ij}$) is subsequently utilized to evaluate the cross-entropy loss with respect to the ground-truth labels. The final semantic clustering loss is defined as
\begin{equation}
    L_{SC} = \sum_{i=1}^{m} \sum_{j=1}^{k} \Upsilon_{ij} \cdot \log\left(\frac{e^{\text{cos-sim}_{ij}}}{\sum_{l=1}^{n} e^{\text{cos-sim}_{il}}}\right).
\end{equation}
Here, $\Upsilon_{ij}$ denotes the one-hot encoding belonging to $i^{th}$ feature proposal and $j^{th}$ class. This approach assists in achieving compact semantic clusters of the known classes and it creates clear separation boundaries between clusters of similar semantics (see Figure~\ref{fig:intro}(a)).

\subsection{Class Decorrelation}
To encourage separation among known clusters, we propose a class decorrelation module inspired by \cite{IDFD}. This module imposes an orthogonality constraint on the proposal features, effectively enhancing the inter-cluster distance, thus facilitating better separation of known classes. Initially, one feature from the set of proposal features $\{F_1, F_2, \ldots, F_m\}$ is sampled for each distinct class within the batch, resulting in a subset of $s$ features $\{F'_1, F'_2, \ldots, F'_s\}$ for $s$ unique classes. Subsequently, we proceed to orthogonalize this subset of features that reduces potential correlations among features, enhancing cluster separation and enabling the model to focus on class-specific differences.

To perform orthogonalization, we first compute the cosine similarity for all the sampled features, forming a similarity matrix. 
\begin{equation}
    \text{sim}_{i,j} = \frac{F'_i \cdot F'_j}{\|F'_i\| \cdot \|F'_j\|}   
\end{equation}
Then we calculate the corresponding correlation between the $i^{th}$ and $j^{th}$ feature (i.e., $\text{corr}_{i,j}$) as given in Equation \ref{eq4}.
\begin{equation} \label{eq4}
    \text{corr}_{i,j} = \frac{e^{\text{sim}_{i,j}}}{\sum_{l=1}^{k} e^{\text{sim}_{i,l}}}
\end{equation} 
Here, $\text{corr}_{i,j}$ gauges both the self-correlation of a feature vector and its dissimilarity from other vectors. The primary objective of this step is to establish inter-dependencies among the chosen features, thereby reflecting the innate properties of various classes present within the feature set. 
Finally, the class decorrelation loss (i.e., $L_{\text{CD}}$) is calculated via cross-entropy between resulting correlation matrix and $s$ unique class-based identity matrix (i.e., $I$).

\begin{equation}\label{eq:CD}
    L_{\text{CD}} = \sum_{i=1}^{s} \sum_{j=1}^{s} I_{i,j} \cdot \log\left(\text{Corr}_{i,j}\right)
\end{equation}
Our objective revolves around diagonalizing this correlation matrix, which enforces decorrelation among the features associated with distinct classes. This not only enhances the clarity of class boundaries but also contributes to improving cluster formation (see Figure~\ref{fig:intro}(a)).


\subsection{Object Focus Loss}
In Faster R-CNN \cite{FasterRCNN}, the RPN places significant emphasis on ground truth objects to acquire knowledge of the objectness score. Nevertheless, this methodology frequently results in the issue of overfitting on the known training classes. This inclination towards overfitting impedes the model's ability to identify new and previously unseen object classes during inference efficiently.

\begin{figure}[t!]
    \centering
        \subfloat[Illustration of centerness score in the calculation of object focus loss. \textcolor{green}{Green colored bounding box} indicate ground-truth annotation while \textcolor{magenta}{magenta colored bounding box} is predicted proposal annotation. \textbf{\textit{d}} denotes the distance between ground-truth annotation and predicted proposal annotation.]{\includegraphics[width=0.99\columnwidth]{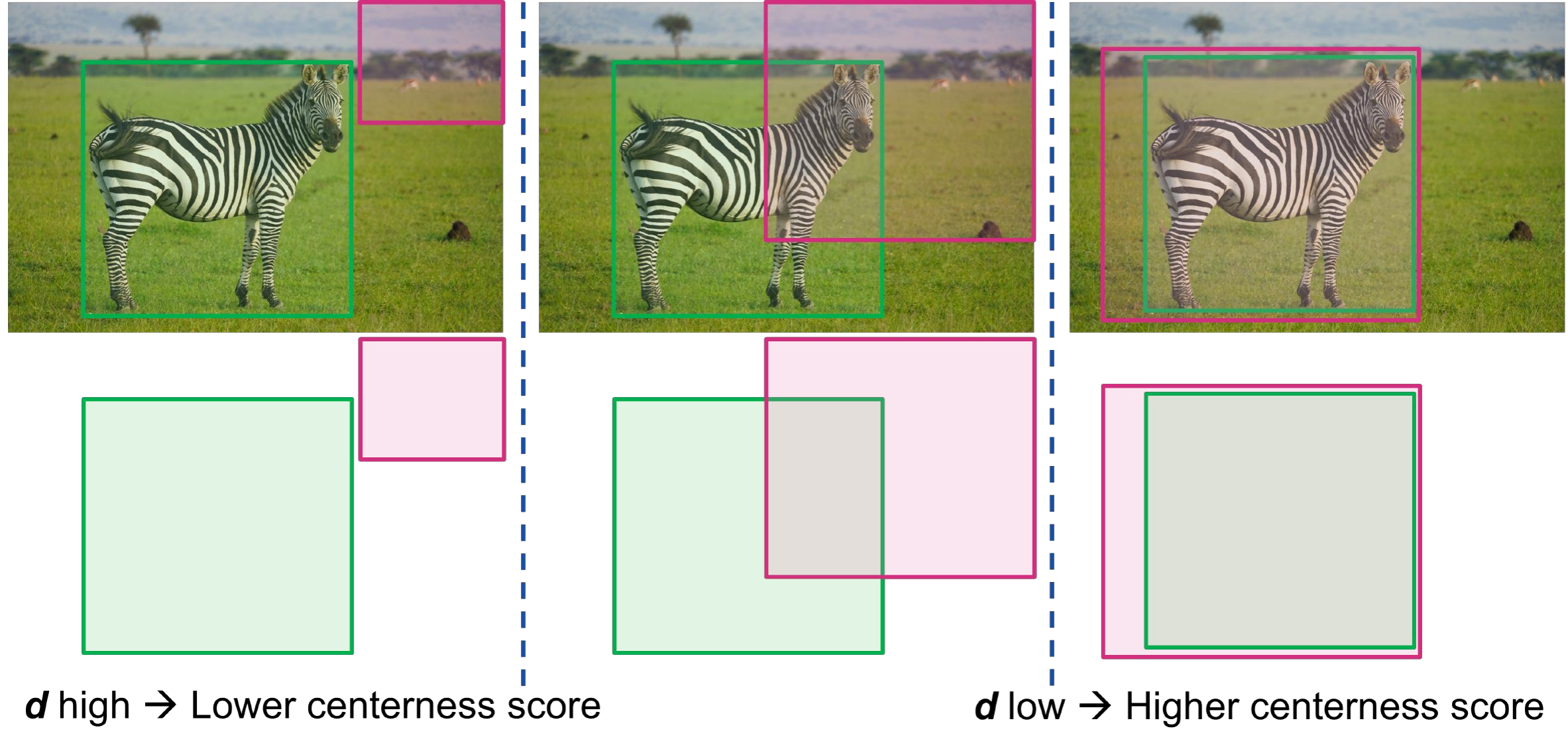}} \\
        \subfloat[Importance of object focus loss in detecting unknown objects]{\includegraphics[width=0.99\columnwidth]{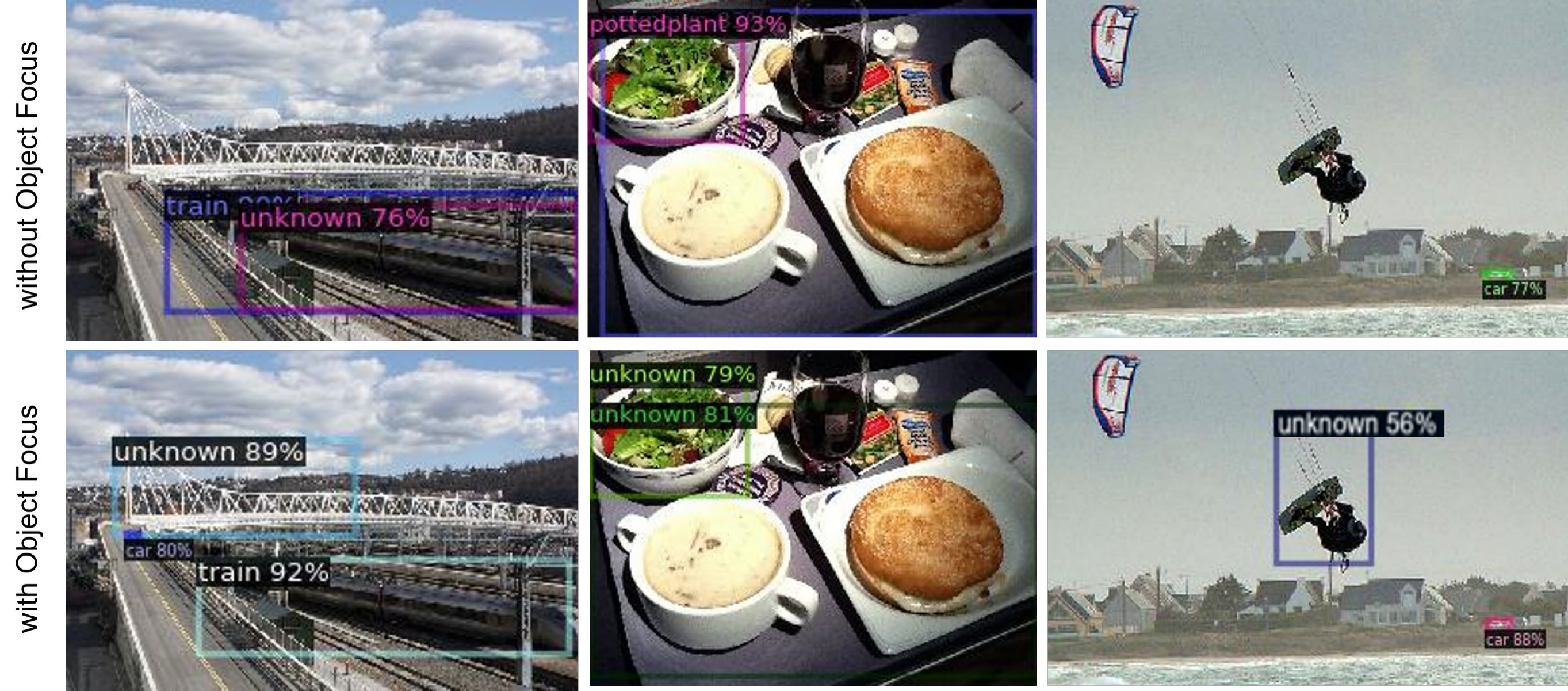}} 
        \vspace{-1em}
\caption{Demonstrates the process of object focus loss along with a visual comparison with and without object focus module.} \label{fig:OF_intro}
\end{figure}
To address this limitation, we introduce object focus loss, which leverages the concept of centerness \cite{centerness,ctrness}. This centerness enables the model to consider critical attributes such as object location, shape, and geometric properties to learn objectness. As a result, we develop a class-agnostic object proposal detection mechanism encompassing a broader range of object characteristics. The centerness loss is computed using centerness logits (i.e., $C_{logits}$) and centerness targets (i.e., $C_{targets}$). In our proposed framework, we have added a single convolution layer after RPN network, which gives us the $C_{logits}$. Furthermore, the corresponding $C_{targets}$ are generated inspired by \cite{centerness, OpenSetRCNN}. It measures how far the center of an object proposal is from the center of a ground-truth bounding box as depicted in Figure~\ref{fig:OF_intro}(a). The details of generating $C_{targets}$ are presented in Supplementary material. The centerness loss (i.e., $L_C$) can be accomplished by evaluating the disparity between the $C_{logits}$ and $C_{targets}$ as given below.
\begin{equation}
   L_C = |C_{logits} - C_{targets}|_1
\end{equation}
Finally, the object focus loss is calculated as the geometric mean of the centerness loss ($L_C$) and the classification-based objectness loss ($L_{Obj}$) \cite{FasterRCNN} as 
\begin{equation}\label{eq:OF}
    L_{Obj-Focus} = \sqrt{L_C \cdot L_{Obj}}.
\end{equation}
The object focus loss (i.e., $L_{Obj-Focus}$) bridges the gap between classification-free and classification-based detection. The introduction of the centerness loss is a powerful remedy, enhancing the model's capability to generalize and detect more objects belonging to unknown classes during evaluation. By shifting the focus towards the geometric aspects of objects, we enable the model to develop a more comprehensive understanding of objects' inherent features.
This results in an enhanced object detection that can effectively handle known and previously unseen classes, improving overall performance and generalization capabilities. We demonstrate its efficacy in Figure~\ref{fig:OF_intro}(b), where we compare results on methods with and without object focus loss. We observe that in selective cases, without object focus loss leads to the failure of detecting certain unknown objects, a problem rectified by incorporating object focus loss.

\subsection{Architectural Details}
We have utilized Faster R-CNN \cite{FasterRCNN} as the baseline network. Further, we add a convolutional layer with a kernel size of $1\times 1$ into the RPN head to enable the regression of centerness logits and the objectness logits generated by the convolutional layer. Moreover, to cluster the $m$ proposal features in the semantic space, two MLP layers are added after the ROI align sampling process. The resultant 1024-dimensional $m$ proposal features are aligned with their corresponding 1024-dimensional text embeddings. The CLIP text encoder \cite{CLIP} is employed to generate the 1024 text embeddings. 
Unlike the CLIP approach \cite{CLIP}, we adopt the class name as a prompt instead of one or more prompts. This setting is validated via ablation analysis presented in Supplementary material.
Furthermore, $m$ proposals are also forwarded to the class decorrelation module, where they are preprocessed to sample one feature from each distinct class in the batch. 


\noindent\textbf{Overall Optimization:}
The proposed framework is trained using final loss $L_{Final}$ i.e., the combination of RPN loss (i.e., $L_{RPN}$) and detector loss (i.e., $L_{Det}$).
\begin{equation}
\begin{aligned}
L_{Final} &= L_{RPN} + L_{Det} \ \ where, \\
    L_{Det} &= \alpha_1 L_{SC} +  \alpha_2 L_{CD} + L_{UPL} + L_{Reg} + L_{CE} \\
    L_{RPN} &= \alpha_3 L_{Obj-Focus} +  L_{RPN-Reg}
\end{aligned}
\end{equation}
Here, $L_{RPN}$ is a weighted combination of object focus loss (i.e., $L_{Obj-Focus}$) and RPN-based regression loss (i.e., $L_{RPN-Reg}$). $L_{Det}$ is a weighted combination of semantic cluster loss (i.e., $L_{SC}$), class decorrelation loss (i.e., $L_{CD}$), unknown probability loss $L_{UPL}$ and standard Faster R-CNN based regression loss (i.e., $L_{Reg}$) and cross-entropy loss (i.e., $L_{CE}$). 
The $L_{UPL}$ is employed from \cite{OpenDet} to learn the unknown probability of a proposal for each instance based on the uncertainty of predictions.

\begin{table*}
\caption{Comparison with SOTA methods on VOC-COCO-T1 setting. The best-performing measures are highlighted with \textbf{bold font}, while the second-best is highlighted with \underline{\textit{underlined italic font}}. * indicates the re-trained methods.} \label{tab:T1}
\vspace{-1em}
\begin{adjustbox}{width=.99\linewidth,center}
\begin{tabular}{|l|c|ccccc|ccccc|ccccc|}
\hline
\multicolumn{1}{|c|}{\multirow{2}{*}{Method}} & VOC    & \multicolumn{5}{c|}{VOC-COCO-20} & \multicolumn{5}{c|}{VOC-COCO-40}  & \multicolumn{5}{c|}{VOC-COCO-60}      \\ \cline{2-17} 
     & $mAP_k \uparrow$ &  $WI \downarrow$      &  $AOSE \downarrow$    &  $mAP_k \uparrow$   & $AP_u \uparrow$ & \textit{HMP} $\uparrow$  &  $WI \downarrow$      &  $AOSE \downarrow$    &  $mAP_k \uparrow$    & $AP_u \uparrow$ & \textit{HMP} $\uparrow$  &  $WI \downarrow$      &  $AOSE \downarrow$    &  $mAP_k \uparrow$    & $AP_u \uparrow$ & \textit{HMP} $\uparrow$ \\ \hline
\multicolumn{17}{|l|}{\textbf{ResNet50 as Backbone}} \\ \hline
Faster R-CNN \cite{FasterRCNN}                                  & \underline{\textit{80.10}}  &  18.39   &  15118   &  58.45    & 0.00 & 0.00    &  22.74   &  23391   &  55.26    & 0.00   & 0.00  &  18.49   &  25472   &  55.83    & 0.00  & 0.00   \\ \hline
ORE \cite{ORE}                                          & 79.80  &  18.18   &  12811   &  58.25    & 2.60  & 4.98 &  22.40   &  19752   &  55.30    & 1.70 & 3.30 &  18.35   &  21415   &  55.47    & 0.53 & 1.05 \\ \hline
DS \cite{DS}                                           & 80.04  &  16.98   &  12868   &  58.35    & 5.13 & 9.43 &  20.86   &  19775   &  55.31    & 3.39 & 6.39 &  17.22   &  21921   &  55.77    & 1.25 & 2.45 \\ \hline
PROSER \cite{PROSER}                                    & 79.68  &  19.16   &  13035   &  57.66    & 10.92 & 18.36 &  24.15   &  19831   &  54.66    & 7.62  & 13.38 &  19.64   &  21322   &  55.20    & 3.25 & 6.14 \\ \hline
OpenDet \cite{OpenDet}                                      & 80.02  &  14.95   &  11286   &  \underline{\textit{58.75}}    & \underline{\textit{14.93}} & \underline{\textit{23.81}} & 18.23   &  16800   &  55.83    & \underline{\textit{10.58}} & \underline{\textit{17.79}} &  14.24   &  \textbf{18250  } &  56.37    & \underline{4.36} & \underline{\textit{8.09}} \\ \hline
Openset RCNN \cite{OpenSetRCNN}                                 & \textbf{82.94}  &  \underline{\textit{11.58}}   &  \underline{\textit{10839}}   &  \textbf{59.19  }  & --- & ---  &  \underline{\textit{14.48}}   &  \underline{\textit{16652}}   &  \textbf{56.12}  & --- & ---  &  \underline{\textit{12.41}}   &  19631   &  \textbf{57.01  }  & --- & ---   \\ \hline
\rowcolor{LightCyan} \textbf{Our (proposed)}                              & 78.06  &  \textbf{9.55}  &  \textbf{9267}  &  58.52  & \textbf{18.45} & \textbf{28.05} & \textbf{11.89} &  \textbf{14057} &  \underline{56.10}  & \textbf{12.56} & \textbf{20.52} &  \textbf{10.96} &  \underline{\textit{19153}} &  \underline{\textit{56.47}}  & \textbf{5.10} & \textbf{9.36} \\ \hline
\multicolumn{17}{|l|}{\textbf{ConvNet-small as Backbone}} \\ \hline
Faster R-CNN* \cite{FasterRCNN}                                  & \textbf{84.88}  & 13.31  & 16019 & \underline{\textit{63.85}}  & 0.00 & 0.00 &   17.15  & 24444    & \underline{\textit{61.25}}   & 0.00 & 0.00   & 13.68 & 26077   & \underline{\textit{62.18}}   & 0.00 & 0.00     \\ \hline
DS* \cite{DS}                                          & 83.24  & 11.22 & 15257 & 62.80  & 6.75 & 12.19  & 14.62   & 23406 & 60.24 & 4.81 & 8.91 & 12.51 & 27951 & 60.58  & 1.79 & 3.48 \\ \hline
PROSER* \cite{PROSER}                   &  \underline{\textit{84.78}} & 16.32 & 14222 & 62.72  & \underline{\textit{18.06}} & \underline{\textit{28.04}} & 20.39  & 21296 & 60.20 & 11.97 & 19.97 & 15.40 & 21889 & 61.03  & 4.57 & 8.50  \\ \hline
OpenDet* \cite{OpenDet}                     & 83.22  &  \underline{\textit{9.43}}  &  \underline{\textit{11700}} &  63.22 & 17.49 & 27.40 &  \underline{\textit{11.79}} &  \underline{\textit{16823}} &  60.83  & \underline{\textit{12.28}} & \underline{\textit{20.43}} &  \underline{\textit{9.61}}  &  \underline{\textit{19963}} & 61.98  & \underline{\textit{4.72}} & \underline{\textit{8.77}} \\ \hline
\rowcolor{LightCyan} \textbf{Our (proposed)}                         & 83.28  &  \textbf{9.06}  &  \textbf{9441} &  \textbf{64.15}  & \textbf{20.60} & \textbf{31.19} &  \textbf{11.20} &  \textbf{13695} &  \textbf{61.56}  & \textbf{14.00} & \textbf{22.81} &  \textbf{9.46}  &  \textbf{17561} &  \textbf{62.26}  & \textbf{5.34} & \textbf{9.84} \\ \hline
\end{tabular}
\end{adjustbox}
\end{table*}
\section{Experiments \& Result Analysis}
\label{sec:results}
\subsection{Implementation details}
\noindent\textbf{Dataset details:} Following \cite{OpenDet}, we use the widely-used PASCAL VOC \cite{PascalVOC} and MS COCO \cite{MSCOCO} benchmark datasets for training and testing purpose. The trainval set of VOC is employed for closed-set training, while 20 VOC classes and 60 non-VOC classes from COCO are adopted to assess the effectiveness of our approach under diverse open-set conditions. Specifically, two settings have been designed, namely VOC-COCO-\{T1, T2\}. 
\begin{itemize}
    \item \textbf{VOC-COCO-T1:} The first setting involves gradually increasing open-set classes to create three joint datasets. Each dataset comprises $n=5000$ VOC testing images, as well as $\{n, 2n, 3n\}$ COCO images with $\{20, 40, 60\}$ non-VOC classes, respectively.
    \item \textbf{VOC-COCO-T2:} This setting involves gradually increasing the Wilderness Ratio (WR) to create four joint datasets. Each dataset comprises $n=5000$ VOC testing images and $\{0.5n, n, 2n, 4n\}$ COCO images that are disjoint with VOC classes. 
\end{itemize} 
\noindent\textbf{Entropy Thresholding:} We employ a technique called entropy thresholding to penalize the low-confidence outputs during evaluation. This involves calculating the entropy of the classification head output logits and comparing them with a fixed threshold. In cases where the entropy ($\mathbf{-p\cdot log(p)}$) of a logit exceeds the threshold, we classify it as the ``unknown" label. This approach ensures that if the open-set detector is not confident in its prediction regarding an object proposal, it is most likely an unknown object that is not encountered during training. After empirical analysis, we have selected a threshold 0.85 for all settings. This thresholding mechanism helps to prevent misclassification of unknown objects as known classes.

\noindent\textbf{Evaluation metrics:} We have performed the evaluation on open-set metrics such as Wilderness Impact ($WI$) \cite{WI}, Absolute Open-Set Error ($AOSE$) \cite{DS}, and $AP_u$ (Average Precision of unknown classes), along with closed-set metric i.e., $mAP_k$ (mean Average Precision of known classes). The purpose of $WI$ \cite{WI} is to determine the degree to which unknown objects have been misclassified into known classes. 
\begin{equation}
    WI = \Big(\frac{P_k}{P_{k\cup u}}-1\Big)\times 100,
\end{equation}
where $P_k$ and $P_{k\cup u}$ denote precision of closed-set and open-set classes, respectively. We report the $WI$ under 0.8 recall level as suggested by \cite{ORE}. Furthermore, $AOSE$ \cite{DS} is utilized to quantify the number of unknown objects that have been misclassified. Lowering the $WI$ and $AOSE$ scores indicates better detection performance. 

\noindent\textbf{Harmonic Mean Precision:} We introduce a new metric called Harmonic Mean Precision (\textit{HMP}) that encapsulates the performance of a detector on both known and unknown classes in one metric. This is achieved by calculating the harmonic mean of the $mAP_k$ and $AP_u$.
\begin{equation}
    HMP = \frac{2 \times mAP_k \times AP_u }{mAP_k + AP_u}
\end{equation}
\vspace{-0.5em}
\begin{table*}
\caption{Comparison with SOTA methods on VOC-COCO-T2 setting. The best-performing measures are highlighted with \textbf{bold font}, while the second-best is highlighted with \underline{\textit{underlined italic font}}. * indicates the re-trained methods.}\label{tab:T2}
\vspace{-1em}
\begin{adjustbox}{width=.99\linewidth,center}
\begin{tabular}{|l|ccccc|ccccc|ccccc|}
\hline
\multirow{2}{*}{Methods} & \multicolumn{5}{c|}{VOC-COCO-n}   & \multicolumn{5}{c|}{VOC-COCO-2n}   & \multicolumn{5}{c|}{VOC-COCO-4n}  \\ \cline{2-16} 
& $WI \downarrow$      &  $AOSE \downarrow$    &  $mAP_k \uparrow$   & $AP_u \uparrow$ & \textit{HMP} $\uparrow$ & $WI \downarrow$      &  $AOSE \downarrow$    &  $mAP_k \uparrow$   & $AP_u \uparrow$ & \textit{HMP} $\uparrow$ & $WI \downarrow$      &  $AOSE \downarrow$    &  $mAP_k \uparrow$   & $AP_u \uparrow$ & \textit{HMP} $\uparrow$ \\ \hline
\multicolumn{16}{|l|}{\textbf{ResNet50 as Backbone}} \\ \hline
Faster R-CNN \cite{FasterRCNN}              & 16.14 & 12409 & 74.52  & 0.00 & 0.00 & 24.18   & 24636   & \underline{70.07}  & 0.00 & 0.00 & 32.89 & 48618 & 63.92  & 0.00 & 0.00 \\ \hline

ORE \cite{ORE}                      & 15.36 & 10568 & 74.34  & 1.80 & 3.51  & 23.67  & 20839   & 70.01 & 2.13 & 4.13 & 32.40 & 40865 & 64.59  & 2.14 & 4.14 \\ \hline
DS \cite{DS}                      & 15.43 & 10136 & 73.67  & 4.11 & 7.79 & 23.21   & 20018 & 69.33  & 4.84 & 9.05 & 31.79 & 39388 & 63.12  & 5.64 & 10.35 \\ \hline
PROSER \cite{PROSER}              & 16.65 & 10601 & 73.55  & 8.88  & 15.85 & 25.74   & 21107   & 69.32  & 10.31 & 17.95 & 34.60 & 41569 & 63.09  & 11.15 & 18.95 \\ \hline
OpenDet \cite{OpenDet}            & 11.70 & 8282  & \textbf{75.56}  & \underline{\textit{12.30}} & \underline{\textit{21.16}}  & \underline{\textit{18.69}} & \underline{\textit{16329}} & \textbf{71.44}  & \underline{\textit{14.96}} & \underline{\textit{24.74}} & 26.69 & \underline{\textit{32419}} & \textbf{65.55}  & \underline{\textit{16.76}} & \underline{\textit{26.69}}  \\ \hline
Openset RCNN \cite{OpenSetRCNN}     & \underline{\textit{11.59}} & \underline{\textit{7705}}  & \underline{\textit{74.90}}  & --- & ---  & ---   & ---   & ---    & ---  & --- & \underline{\textit{25.14}} & 34382 & \underline{\textit{64.88}}  & --- & ---     \\ \hline
\rowcolor{LightCyan} \textbf{Our (proposed)}         & \textbf{9.51}   &  \textbf{6875}   &  73.47   & \textbf{14.13} & \textbf{23.70} &  \textbf{15.39}  &  \textbf{13615}  &  69.44   & \textbf{16.36} & \textbf{26.48} &  \textbf{22.31}  &  \textbf{27362}  &  64.28   & \textbf{17.77} & \textbf{27.84}  \\ \hline
\multicolumn{16}{|l|}{\textbf{ConvNet-small as Backbone}} \\ \hline 
Faster R-CNN* \cite{FasterRCNN}         & 14.23 & 11611 & 79.04  & 0.00  & 0.00 & 23.47 & 23190  & 73.63  & 0.00 & 0.00 & 33.52 & 46418 & 67.11  & 0.00 & 0.00 \\ \hline
DS* \cite{DS}                & 14.18 & 11166 & 76.29  & 5.15 & 9.65  & 22.85  & 22200  & 71.39  & 6.46 & 11.85 & 32.84 & 44648 & 64.98  & 7.69 & 13.75 \\ \hline
PROSER* \cite{PROSER}     & 13.35 & 10053 & 78.04  & 12.85 & 22.07 & 22.01 &  20255 & 72.63 & 15.07 & 24.96 & 32.39 & 40389 & 65.98  & 17.02 & 27.06\\ \hline
OpenDet* \cite{OpenDet}   &  \textbf{9.05}   &  \underline{\textit{7615}}   &  \underline{\textit{79.27}}   & \underline{\textit{13.79}} & \underline{\textit{23.49}}  &  \underline{\textit{15.80}}  &  \underline{\textit{15150}}  &  \underline{\textit{74.79}}   & \underline{\textit{16.64}}  &  \underline{\textit{27.22}} & \underline{\textit{24.53}}  &  \underline{\textit{30159}}  &  \underline{\textit{69.20}}   & \underline{\textit{18.74}}  & \underline{\textit{29.49}}                    \\ \hline
\rowcolor{LightCyan} \textbf{Our (proposed)}  &  \underline{\textit{9.43}}   &  \textbf{6969}   &  \textbf{79.69}   & \textbf{15.33}    &  \textbf{25.71} & \textbf{15.60}  &  \textbf{11977}  &  \textbf{75.74}   & \textbf{17.87} & \textbf{28.92} & \textbf{24.29}  &  \textbf{27654}  &  \textbf{70.27}   & \textbf{19.47} & \textbf{30.49}  \\ \hline
\end{tabular}
\end{adjustbox}
\end{table*}
\begin{figure*}[t!]
    \centering
    \includegraphics[width = 0.99\linewidth]{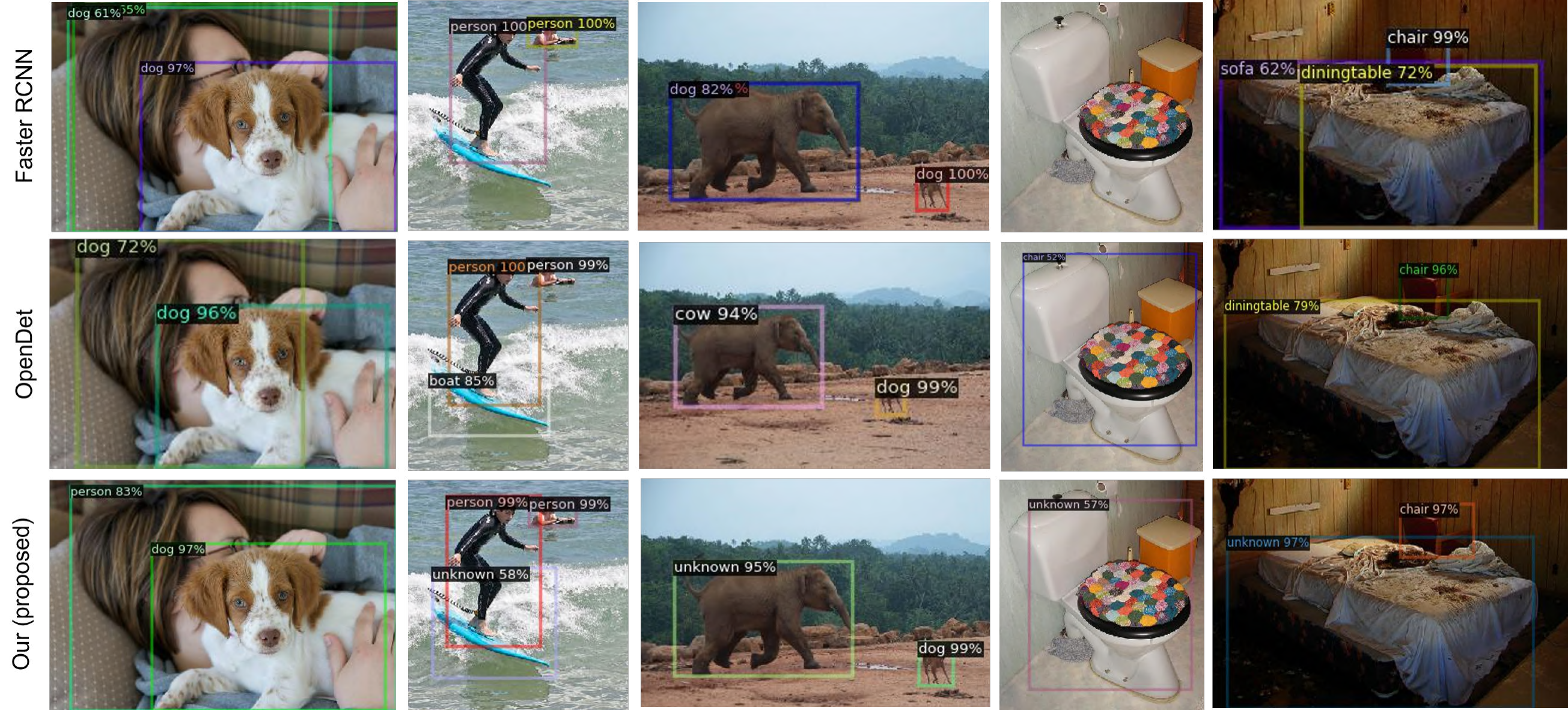}
    \vspace{-1em}
    \caption{\small Visual comparison between our proposed model and baseline methods such as Faster R-CNN \cite{FasterRCNN} and OpenDet \cite{OpenDet}. More results can be visualized from the Supplementary materials. (Zoom-in for a better view)}
    \label{fig:visualize}
    \vspace{-1em}
\end{figure*}

\noindent\textbf{Training details:}
For a fair comparison with existing OSOD methods, we have adopted the Faster R-CNN \cite{FasterRCNN} architecture as our baseline and incorporated our modules to transform it into an open-set detector. During the training phase, for experiments with the ConvNet backbone, we utilized the AdamW optimizer with a learning rate of 1e-4 and trained for 50,000 iterations. Our training process has involved using a single GPU with a batch size of 6. In the case of experiments with the ResNet50 backbone, we opted for the SGD optimizer with a learning rate of 0.002 and trained for 32,000 iterations as well. Like the ConvNet experiments, the training has been conducted using a single GPU but with a larger batch size of 16. 
We empirically set the weight coefficients $\alpha_1$ and $\alpha_2$ to 0.05, and $\alpha_3$ to 1. 
For comparison, we choose recent OSOD methods such as Faster R-CNN \cite{FasterRCNN}, ORE \cite{ORE}, DS \cite{DS}, PROSER \cite{PROSER}, OpenDet \cite{OpenDet}, and Openset RCNN \cite{OpenSetRCNN}. These methods are re-trained on the exact configuration to compare the results on the CovnNet backbone. Furthermore, we have also compared our method with open world object detection methods such as ORE \cite{ORE}, OW-DETR \cite{OW-DETR}, PROB \cite{PROB} on OSOD-based evaluation protocol proposed in \cite{ORE} and the corresponding results are presented in Supplementary material.

\subsection{Result Analysis}
\noindent\textbf{Comparison on T1 setting:}
Table \ref{tab:T1} presents a comparison of state-of-the-art (SOTA) methods on VOC-COCO-T1 setting. This comparison is conducted by utilizing the ResNet50 and ConvNet backbone in terms of open-set and closed-set metrics. The results from Table \ref{tab:T1} indicate that, in the context of ResNet50 backbone comparison, the proposed method improves $AP_u$  by 17-24\% from OpenDet \cite{OpenDet} all dataset settings. Our method also reduces $AOSE$ by 450-2600 and gains 11-18\% in $WI$ compared to previous best-performing Openset RCNN \cite{OpenSetRCNN} method. For the open-set $mAP_k$ measure, the proposed model performs better than the OpenDet \cite{OpenDet} method on all the settings except VOC-COCO-20. When comparing the results based on the ConvNet backbone, the proposed framework outperforms other methods across all metrics, achieving an increase by 13-18\% in $AP_u$ and reducing $AOSE$ by approximately 2200-3200 compared to the previous best-performing method \cite{OpenDet}. We also show a gain in $HMP$ by 11-18\% across all settings on both backbones against \cite{OpenDet}.

\noindent\textbf{Comparison on T2 setting:} 
In Table \ref{tab:T2}, a comparison between the proposed and existing SOTA methods on the VOC-COCO-T2 setting is presented. In the context of ResNet50-based comparison, the proposed model outperforms other methods in terms of $WI$, $AOSE$, and $AP_u$ across all dataset settings. For example in VOC-COCO-n, we show an increase of 14.9\% in $AP_u$ from \cite{OpenDet}.
Similarly, in comparison on the ConvNet backbone, the proposed method outperforms other methods on all dataset settings, except on the $WI$ measure in the VOC-COCO-n setting, where the proposed method achieves a comparable measure with OpenDet \cite{OpenDet}. With ConvNet backbone, our method achieves a 4-12\% increase in $AP_u$, and consistently improves upon $AOSE$ by a large margin of 600-3200 compared to \cite{OpenDet}. In $HMP$, we improve upon \cite{OpenDet} by 3-12\% across all settings on both backbones.

\noindent \textbf{Visual Comparison:}
In addition to quantitative analysis, a qualitative comparison is provided in Figure \ref{fig:visualize} to demonstrate the improvement of our method over baseline methods, i.e., Faster R-CNN \cite{FasterRCNN}) and previous best-performing OpenDet \cite{OpenDet}.
It can be visualized that the proposed method accurately classifies unknown objects that are semantically closer to known classes which other methods fails to do.
For example, OpenDet \cite{OpenDet} misclassifies `bed' as `dining table' due to their semantic similarity. However, our model, having learned semantic-based clusters, correctly labels `bed' as the `unknown' class. A similar analysis is conducted on `elephant' and `toilet', which OpenDet \cite{OpenDet} misclassifies as `cow' and `chair', respectively. Our method, however, accurately identifies these as `unknown' classes. 

\subsection{Ablation Studies \& Analysis}
\label{sec:ablation}
To ensure a fair comparison, all ablation experiments were trained using ConvNet backbone and evaluated on the VOC-COCO-40 setting.

\noindent\textbf{Effects of proposed method's components:} 
In this investigation, we examine the impact of each component of the proposed framework, i.e., semantic clustering (\textit{SC}), class decorrelation (\textit{CD}) and object focus (\textit{OF})\footnote{without object focus (\textit{OF}) refers to only standard RPN based objectness loss $L_{Obj}$}. The proposed framework is trained utilizing either individual or combined introduced components to assess their efficacy on the comprehensive performance. The corresponding outcomes are presented in Table \ref{tab:abl1}, which reveals that all three introduced components significantly improve the performance of the proposed framework in known and unknown evaluation metrics. Furthermore, one can notice that adding each proposed component improves the performance of detector. For instance, adding \textit{OF} module to \textit{SC} and \textit{CD} module (i.e., Case 5 and Case 6) improves the known as well as unknown detection performance, proving the importance of \textit{OF} module, similarly, adding \textit{CD} module to \textit{SC} i.e., Case 4, enhance the detection performance as compared to Case 1 and Case 2. This substantiates the consequence of our introduced modules in the proposed framework.
\begin{table}[t!]
\caption{Ablation analysis to validate the proposed components: semantic clustering (\textit{SC}), class decorrelation (\textit{CD}) and Object Focus loss (\textit{OF}) on VOC-COCO-40 setting.}
\label{tab:abl1}
\vspace{-1em}
\begin{adjustbox}{width=.99\linewidth,center}
\begin{tabular}{|c|ccc|ccccc|}
\hline
  & \textit{SC} & \textit{CD} & \textit{OF} & $WI \downarrow$      &  $AOSE \downarrow$    &  $mAP_k \uparrow$   & $AP_u \uparrow$ & $HMP \uparrow$  \\ \hline
Case 1                 & \checkmark  &    &   & 11.47 & 15561  & 60.22   & 12.29 &    20.41  \\ \hline
Case 2                 &    & \checkmark  &    & 12.26   & 16041  & 59.58   & 11.84 &  19.75    \\ \hline
Case 3                 &    &    & \checkmark  &  11.91  &   15268   &   60.34    &   12.22 & 20.32 \\ \hline
Case 4                 & \checkmark  & \checkmark  &    & 11.28   &   16301 &  61.12     &  12.31 &  20.49 \\ \hline
Case 5                 & \checkmark  &    & \checkmark  &  11.30  &   15225   &  61.52     &   13.59 & 22.26 \\ \hline
Case 6                 &    & \checkmark  & \checkmark  &  12.09  &   15642   &   60.42    &  13.95  &  22.67\\ \hline
\rowcolor{LightCyan} Proposed                 & \checkmark  & \checkmark  & \checkmark  &  \textbf{11.20}  &   \textbf{13695}   &   \textbf{61.56}    &   \textbf{14.00} &  \textbf{22.81} \\ \hline
\end{tabular}
\end{adjustbox}
\end{table}

\noindent\textbf{Effects of geometric mean operation in Object Focus loss:}
In proposed object focus loss, we have employed the geometric mean between $L_{C}$ and $L_{Obj}$ (see Equation \ref{eq:OF}). To validate of this operation, we have conducted a series of experiments utilizing various settings, including only RPN-based object loss (i.e., $L_{Obj}$), only centerness loss (i.e., $L_C$), as well as the addition and multiplication of $L_{Obj}$ and $L_C$. The results of these experiments are presented in Table \ref{tab:GM-OF}, which demonstrate that the combination of $L_C$ and $L_{Obj}$ through the geometric mean operation performs better than the other settings.
\begin{table}[t!]
\caption{Ablation analysis to validate the geometric mean operation in object focus loss on VOC-COCO-40 setting.} \label{tab:GM-OF}
\vspace{-1em}
\begin{adjustbox}{width=.99\linewidth,center}
\begin{tabular}{|l|ccccc|}
\hline
  & $WI \downarrow$      &  $AOSE \downarrow$    &  $mAP_k \uparrow$   & $AP_u \uparrow$   & $HMP \uparrow$\\ \hline
Only $L_{Obj}$     &  11.28  & 16301  & 61.12  &  12.31 & 20.49\\ \hline
Only $L_C$      &  14.53  & 6758  & 15.86   & 2.78 & 4.73\\ \hline
$L_{Obj} + L_C$    &  11.97  & 15260     &  60.36   & 11.87 &  19.84  \\ \hline
$L_{Obj} \times L_C$  &  \textbf{11.14}  &  19645  & 43.70   & 10.67 & 17.15 \\ \hline
\rowcolor{LightCyan} $\sqrt{L_{Obj}\cdot L_C}$ (proposed) &  11.20  & \textbf{13695}  &  \textbf{61.56}  & \textbf{14.00} & \textbf{22.81} \\ \hline
\end{tabular}
\end{adjustbox}
\end{table}
\begin{figure}[t!]
    \centering
    \includegraphics[width = 0.99\linewidth]{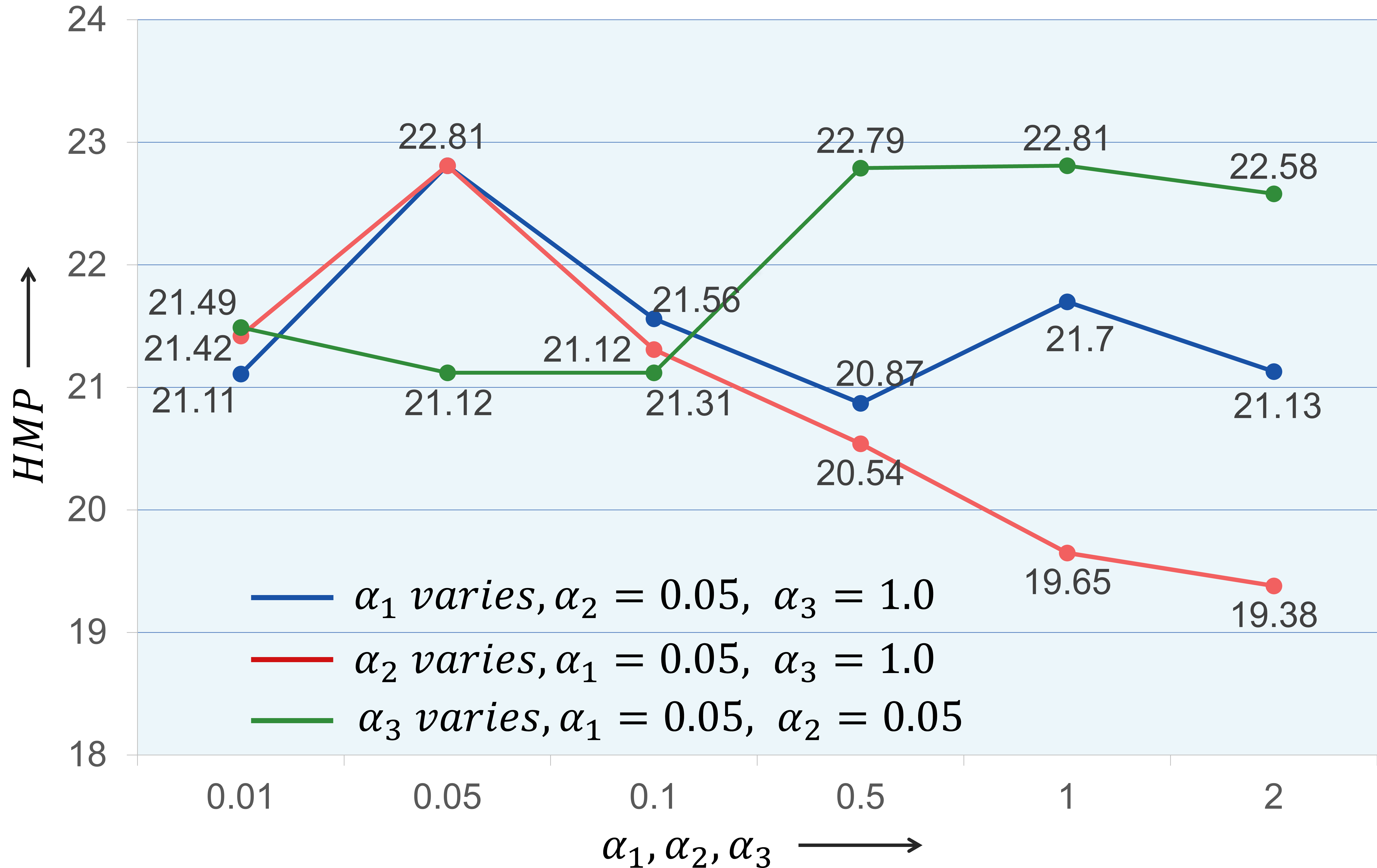}
    \vspace{-1em}
    \caption{\small Effect of weight coefficients $\alpha_1$, $\alpha_2$ and $\alpha_3$ in terms of \textit{HMP} measure on VOC-COCO-40 setting.}
    \label{fig:weight-Abl}
    \vspace{-1em}
\end{figure}

\noindent\textbf{Effect of loss weights:} 
The proposed framework is trained using a weighted combination of multi-task losses, wherein the weight coefficients for $L_{SC}$, $L_{CD}$ and $L_{Obj-Focus}$ are represented by $\alpha_1$, $\alpha_2$ and $\alpha_3$, respectively. To determine these weight coefficients, a few experiments were conducted and the corresponding results in terms of \textit{HMP} measure are illustrated in Figure~\ref{fig:weight-Abl}. The graph portrays the analysis accomplished by varying one weight coefficient while keeping the value of the remaining coefficients fixed. After conducting empirical analysis, it was discovered that the $\alpha_1 = 0.05$, $\alpha_2 = 0.05$ and $\alpha_3 = 1.0$ combinations give us better \textit{HMP} measure than other settings.

\section{Conclusion \& Future Work} 
\label{sec:conclusion}
This paper proposes a new framework that effectively aligns known class representations to detect unknown objects accurately. The proposed model offers a solution to the issue of high-confidence unknown misclassification in OSOD. We demonstrate that clustering in the semantic space facilitates the formation of well-defined boundaries between clusters, particularly for semantically similar classes that are highly susceptible to misclassification. Additionally, we introduce a class decorrelation module that promotes inter-cluster separation and an object focus loss, wherein the objectness learning exhibits robustness in detecting novel and unseen objects. We also employ an entropy-thresholding-based evaluation technique that penalizes low-confidence outputs, thereby reducing the risk of misclassifying unknown objects. Finally, we carried out extensive experiments \& ablation studies and found that the proposed method outperforms existing SOTA methods with significant margin. 
As the proposed approach have great potential to improve class alignment, it can be further extended to other open-set tasks like incremental object detection and open-set domain adaptation.
\balance
{\small
\bibliographystyle{ieee_fullname}
\bibliography{egbib}

\begin{thebibliography}{10}\itemsep=-1pt

\bibitem{OSR5}
Abhijit Bendale and Terrance~E Boult.
\newblock Towards open set deep networks.
\newblock In {\em Proceedings of the IEEE conference on computer vision and pattern recognition}, pages 1563--1572, 2016.

\bibitem{WI}
Akshay Dhamija, Manuel Gunther, Jonathan Ventura, and Terrance Boult.
\newblock The overlooked elephant of object detection: Open set.
\newblock In {\em Proceedings of the IEEE/CVF Winter Conference on Applications of Computer Vision}, pages 1021--1030, 2020.

\bibitem{PascalVOC}
Mark Everingham, Luc Van~Gool, Christopher~KI Williams, John Winn, and Andrew Zisserman.
\newblock The pascal visual object classes (voc) challenge.
\newblock {\em International journal of computer vision}, 88:303--338, 2010.

\bibitem{OSR6}
ZongYuan Ge, Sergey Demyanov, Zetao Chen, and Rahil Garnavi.
\newblock Generative openmax for multi-class open set classification.
\newblock {\em arXiv preprint arXiv:1707.07418}, 2017.

\bibitem{OW-DETR}
Akshita Gupta, Sanath Narayan, KJ Joseph, Salman Khan, Fahad~Shahbaz Khan, and Mubarak Shah.
\newblock Ow-detr: Open-world detection transformer.
\newblock In {\em Proceedings of the IEEE/CVF Conference on Computer Vision and Pattern Recognition}, pages 9235--9244, 2022.

\bibitem{OpenDet}
Jiaming Han, Yuqiang Ren, Jian Ding, Xingjia Pan, Ke Yan, and Gui-Song Xia.
\newblock Expanding low-density latent regions for open-set object detection.
\newblock In {\em Proceedings of the IEEE/CVF Conference on Computer Vision and Pattern Recognition}, pages 9591--9600, 2022.

\bibitem{CC1}
Kaiming He, Haoqi Fan, Yuxin Wu, Saining Xie, and Ross Girshick.
\newblock Momentum contrast for unsupervised visual representation learning.
\newblock In {\em Proceedings of the IEEE/CVF conference on computer vision and pattern recognition}, pages 9729--9738, 2020.

\bibitem{ORE}
KJ Joseph, Salman Khan, Fahad~Shahbaz Khan, and Vineeth~N Balasubramanian.
\newblock Towards open world object detection.
\newblock In {\em Proceedings of the IEEE/CVF conference on computer vision and pattern recognition}, pages 5830--5840, 2021.

\bibitem{CC3}
Prannay Khosla, Piotr Teterwak, Chen Wang, Aaron Sarna, Yonglong Tian, Phillip Isola, Aaron Maschinot, Ce Liu, and Dilip Krishnan.
\newblock Supervised contrastive learning.
\newblock {\em Advances in neural information processing systems}, 33:18661--18673, 2020.

\bibitem{ctrness}
Dahun Kim, Tsung-Yi Lin, Anelia Angelova, In~So Kweon, and Weicheng Kuo.
\newblock Learning open-world object proposals without learning to classify.
\newblock {\em IEEE Robotics and Automation Letters}, 7(2):5453--5460, 2022.

\bibitem{MSCOCO}
Tsung-Yi Lin, Michael Maire, Serge Belongie, James Hays, Pietro Perona, Deva Ramanan, Piotr Doll{\'a}r, and C~Lawrence Zitnick.
\newblock Microsoft coco: Common objects in context.
\newblock In {\em Computer Vision--ECCV 2014: 13th European Conference, Zurich, Switzerland, September 6-12, 2014, Proceedings, Part V 13}, pages 740--755. Springer, 2014.

\bibitem{SwinT}
Ze Liu, Yutong Lin, Yue Cao, Han Hu, Yixuan Wei, Zheng Zhang, Stephen Lin, and Baining Guo.
\newblock Swin transformer: Hierarchical vision transformer using shifted windows.
\newblock In {\em Proceedings of the IEEE/CVF international conference on computer vision}, pages 10012--10022, 2021.

\bibitem{DS}
Dimity Miller, Lachlan Nicholson, Feras Dayoub, and Niko S{\"u}nderhauf.
\newblock Dropout sampling for robust object detection in open-set conditions.
\newblock In {\em 2018 IEEE International Conference on Robotics and Automation (ICRA)}, pages 3243--3249. IEEE, 2018.

\bibitem{OSR7}
Lawrence Neal, Matthew Olson, Xiaoli Fern, Weng-Keen Wong, and Fuxin Li.
\newblock Open set learning with counterfactual images.
\newblock In {\em Proceedings of the European Conference on Computer Vision (ECCV)}, pages 613--628, 2018.

\bibitem{OSR8}
Poojan Oza and Vishal~M Patel.
\newblock C2ae: Class conditioned auto-encoder for open-set recognition.
\newblock In {\em Proceedings of the IEEE/CVF Conference on Computer Vision and Pattern Recognition}, pages 2307--2316, 2019.

\bibitem{CC2}
Rui Qian, Tianjian Meng, Boqing Gong, Ming-Hsuan Yang, Huisheng Wang, Serge Belongie, and Yin Cui.
\newblock Spatiotemporal contrastive video representation learning.
\newblock In {\em Proceedings of the IEEE/CVF Conference on Computer Vision and Pattern Recognition}, pages 6964--6974, 2021.

\bibitem{CLIP}
Alec Radford, Jong~Wook Kim, Chris Hallacy, Aditya Ramesh, Gabriel Goh, Sandhini Agarwal, Girish Sastry, Amanda Askell, Pamela Mishkin, Jack Clark, et~al.
\newblock Learning transferable visual models from natural language supervision.
\newblock In {\em International conference on machine learning}, pages 8748--8763. PMLR, 2021.

\bibitem{FasterRCNN}
Shaoqing Ren, Kaiming He, Ross Girshick, and Jian Sun.
\newblock Faster r-cnn: Towards real-time object detection with region proposal networks.
\newblock In {\em Advances in Neural Information Processing Systems}, volume~28, 2015.

\bibitem{OSR9}
Bo Sun, Banghuai Li, Shengcai Cai, Ye Yuan, and Chi Zhang.
\newblock Fsce: Few-shot object detection via contrastive proposal encoding.
\newblock In {\em Proceedings of the IEEE/CVF Conference on Computer Vision and Pattern Recognition}, pages 7352--7362, 2021.

\bibitem{CC4}
Bo Sun, Banghuai Li, Shengcai Cai, Ye Yuan, and Chi Zhang.
\newblock Fsce: Few-shot object detection via contrastive proposal encoding.
\newblock In {\em Proceedings of the IEEE/CVF Conference on Computer Vision and Pattern Recognition}, pages 7352--7362, 2021.

\bibitem{centerness}
Zhi Tian, Chunhua Shen, Hao Chen, and Tong He.
\newblock Fcos: Fully convolutional one-stage object detection.
\newblock In {\em Proceedings of the IEEE/CVF international conference on computer vision}, pages 9627--9636, 2019.

\bibitem{OWOD2}
Yan Wu, Xiaowei Zhao, Yuqing Ma, Duorui Wang, and Xianglong Liu.
\newblock Two-branch objectness-centric open world detection.
\newblock In {\em Proceedings of the 3rd International Workshop on Human-Centric Multimedia Analysis}, pages 35--40, 2022.

\bibitem{ctrness_rel}
Yan Wu, Xiaowei Zhao, Yuqing Ma, Duorui Wang, and Xianglong Liu.
\newblock Two-branch objectness-centric open world detection.
\newblock In {\em Proceedings of the 3rd International Workshop on Human-Centric Multimedia Analysis}, pages 35--40, 2022.

\bibitem{UC-OWOD}
Zhiheng Wu, Yue Lu, Xingyu Chen, Zhengxing Wu, Liwen Kang, and Junzhi Yu.
\newblock Uc-owod: Unknown-classified open world object detection.
\newblock In {\em European Conference on Computer Vision}, pages 193--210. Springer, 2022.

\bibitem{IDFD}
Kouta~Nakata Yaling~Tao, Kentaro~Takagi.
\newblock Clustering-friendly representation learning via instance discrimination and feature decorrelation.
\newblock {\em Proceedings of ICLR 2021}, 2021.

\bibitem{OSR10}
Ryota Yoshihashi, Wen Shao, Rei Kawakami, Shaodi You, Makoto Iida, and Takeshi Naemura.
\newblock Classification-reconstruction learning for open-set recognition.
\newblock In {\em Proceedings of the IEEE/CVF Conference on Computer Vision and Pattern Recognition}, pages 4016--4025, 2019.

\bibitem{OWOD1}
Jinan Yu, Liyan Ma, Zhenglin Li, Yan Peng, and Shaorong Xie.
\newblock Open-world object detection via discriminative class prototype learning.
\newblock In {\em 2022 IEEE International Conference on Image Processing (ICIP)}, pages 626--630. IEEE, 2022.

\bibitem{CC5}
Xiangyun Zhao, Raviteja Vemulapalli, Philip~Andrew Mansfield, Boqing Gong, Bradley Green, Lior Shapira, and Ying Wu.
\newblock Contrastive learning for label efficient semantic segmentation.
\newblock In {\em Proceedings of the IEEE/CVF International Conference on Computer Vision}, pages 10623--10633, 2021.

\bibitem{PROSER}
Da-Wei Zhou, Han-Jia Ye, and De-Chuan Zhan.
\newblock Learning placeholders for open-set recognition.
\newblock In {\em Proceedings of the IEEE/CVF conference on computer vision and pattern recognition}, pages 4401--4410, 2021.

\bibitem{OpenSetRCNN}
Zhongxiang Zhou, Yifei Yang, Yue Wang, and Rong Xiong.
\newblock Open-set object detection using classification-free object proposal and instance-level contrastive learning.
\newblock {\em IEEE Robotics and Automation Letters}, 8(3):1691--1698, 2023.

\bibitem{DETR}
Xizhou Zhu, Weijie Su, Lewei Lu, Bin Li, Xiaogang Wang, and Jifeng Dai.
\newblock Deformable detr: Deformable transformers for end-to-end object detection.
\newblock {\em arXiv preprint arXiv:2010.04159}, 2020.

\bibitem{PROB}
Orr Zohar, Kuan-Chieh Wang, and Serena Yeung.
\newblock Prob: Probabilistic objectness for open world object detection.
\newblock In {\em Proceedings of the IEEE/CVF Conference on Computer Vision and Pattern Recognition}, pages 11444--11453, 2023.

\end{thebibliography}
}

\newpage
\appendix
\section*{Supplementary Material}


\section{ Experimental Settings}\label{sup:exp_setting}
\subsection{ Dataset Details}
We used the Pascal VOC \cite{PascalVOC} and MS-COCO \cite{MSCOCO} for training and testing purposes.
This section presents more details about these datasets and the open-set object detection (OSOD) based evaluation settings.
\\
\noindent\textbf{Pascal VOC \cite{PascalVOC}:}
It contains VOC07 \textit{trainval} set having $5,011$ images, and VOC12 \textit{trainval} set having $11,540$ images with $20$ labeled classes. Further, VOC07 \textit{val} split set is taken as a validation dataset. \\
\textbf{MS-COCO \cite{MSCOCO}:}
This dataset comprises a training set of more than $118,000$ images with $80$ labeled classes While the validation dataset (\textit{val2017}) contains $5000$ labeled images.

The process of closed-set training is executed on VOC07 \textit{trainval} and VOC12 \textit{trainval} set. While the close-set performance is evaluated on the \textit{test} split of VOC07. For testing under open-set conditions, we follow the evaluation protocol suggested in \cite{OpenDet} where testing images having $20$ VOC classes and $60$ non-VOC classes \cite{MSCOCO} are employed and categorized in two settings named as VOC-COCO-T1 and VOC-COCO-T2. 
\begin{itemize}
    \item \textbf{VOC-COCO-T1:} In this setting, the $80$ COCO classes have been categorized into four groups, each comprising 20 classes, based on their semantics \cite{ORE, OpenDet}. To create VOC-COCO-$\{20, 40, 60\}$, we utilized 5000 VOC testing images and $\{n, 2n, 3n\}$ COCO images, each of which contained $\{20, 40, 60\}$ non-VOC classes with semantic shifts, respectively.
    \item \textbf{VOC-COCO-T2:} In this setting, four datasets have been constructed by gradually increasing the wilderness ratio while utilizing $n=5000$ VOC testing images and $\{0.5n, n, 2n, 4n\}$ COCO images, disjointing with VOC classes. Unlike the VOC-COCO-T1 setting, the VOC-COCO-T1 aims to assess the model's performance under significantly greater wilderness, whereby a substantial quantity of testing instances remain unseen during the training process.
\end{itemize}
\subsection{ Implementation Details}
In addition to experimental analysis on ResNet50 and ConvNet backbone presented in main manuscript, we present further analysis on Swin-T backbone \cite{SwinT}. To do such experiments, we have opted to utilize AdamW as an optimizer with a learning rate of 1e-4 and trained for 32,000 iterations with a 0.05 weight decay rate during training phase. The training process has been facilitated by a single GPU with a batch size of 6. For fair comparison, we re-train the Faster R-CNN \cite{FasterRCNN}, DS \cite{DS}, PROSER \cite{PROSER} and OpenDet \cite{OpenDet} methods on same configuration.
\\
\noindent\textbf{Open World Object Detection (OWOD) setting:}
To demonstration how the proposed method performs in the context of OWOD setting, we conducted evaluation according to the ORE protocol \cite{ORE}\footnote{https://github.com/JosephKJ/OWOD}, which was specifically designed for OWOD and comprises four tasks aimed at assessing the performance of OSOD and incremental learning. However, as our work is not concerned with incremental learning, we restrict our evaluation to task 1. The dataset utilized for task 1 comprises $16551$ Pascal VOC images with $20$ classes \cite{PascalVOC} for training and the $10246$ testing images having $20$ VOC classes and $60$ COCO classes \cite{MSCOCO} for open-set evaluation. Here, we compare the proposed method against the baseline Faster R-CNN \cite{FasterRCNN} and its oracle version\footnote{An 'Oracle' detector is a reference model that has access to all known and unknown labels at any given point \cite{ORE}.}, in addition to OWOD methods (ORE \cite{ORE}, OW-DETR \cite{OW-DETR}, PROB \cite{PROB}) and OSOD methods (OpenDet \cite{OpenDet} and Openset RCNN \cite{OpenSetRCNN}).

\subsection{ Generation of centerness targets}
This section presents the procedure of generating the centerness target, i.e., $C_{targets}$ for calculating the centerness loss. The corresponding steps are mentioned below.
\begin{itemize}
    \item The initial step involves the conversion of the default ground-truth bounding box and proposal coordinates, which are in ($x_1,y_1,x_2,y_2$) format, to $\{cx, cy, h, w\}$ format. This conversion results in the center coordinates represented by $cx$ and $cy$, while $h$ and $w$ represent the height and width of the bounding box or proposal, respectively. In the case of a ground-truth box $i$ and proposal box $j$, the transformed bounding box and proposal targets can be denoted by \{\(\textit{cx}_{\textit{gt}(i)}, \textit{cy}_{\textit{gt}(i)}, \textit{h}_{\textit{gt}(i)}, \textit{w}_{\textit{gt}(i)}\)\} and \{\(\textit{cx}_{\textit{p}(j)}, \textit{cy}_{\textit{p}(j)}, \textit{h}_{\textit{p}(j)}, \textit{w}_{\textit{p}(j)}\)\}, respectively.
    \item Subsequently, the differences in those quantities between the proposal box $j$ and the ground truth boxes are determined. Concerning the ground-truth box $i$, the differences can be computed as follows.
\begin{equation*}
    dx_{ij} = \frac{\textit{cx}_{\textit{gt}(i)} - \textit{cx}_{\textit{p}(j)}}{\textit{w}_{\textit{p}(j)}}
\end{equation*}
\begin{equation*}
    dy_{ij} = \frac{\textit{cy}_{\textit{gt}(i)} - \textit{cy}_{\textit{p}(j)}}{\textit{h}_{\textit{p}(j)}}
\end{equation*}
\begin{equation*}
    dw_{ij} = \log\left(\frac{\textit{w}_{\textit{gt}(i)}}{\textit{w}_{\textit{p}(j)}}\right)
\end{equation*}
\begin{equation*}
    dh_{ij} = \log\left(\frac{\textit{h}_{\textit{gt}(i)}}{\textit{h}_{\textit{p}(j)}}\right)
\end{equation*}
    Here, we filter out the targets with negative values.
    Finally, the centerness target for proposal box $j$ and ground-truth box $i$ can be calculated as given in \cite{centerness}.
\begin{equation*}
    C_{target} = \sqrt{\frac{min(dx_{ij},dy_{ij})}{max(dx_{ij},dy_{ij})} \cdot \frac{min(dw_{ij},dh_{ij})}{max(dw_{ij},dh_{ij})}},
\end{equation*}
where, $min(\cdot)$ and $max(\cdot)$ denote the minimum and maximum operations. 
\end{itemize}

\section{ Ablation Studies \& Analysis}\label{sup:Ablation}
This section presents additional ablation analysis to establish the efficacy of the proposed framework. All ablation experiments are trained using ConvNet backbone to ensure a fair comparison and evaluated on the VOC-COCO-40 setting.
\subsection{Effect of prompts in Semantic Clustering module}
\begin{figure}[h!]
    \centering
    \includegraphics[width = 0.99\linewidth]{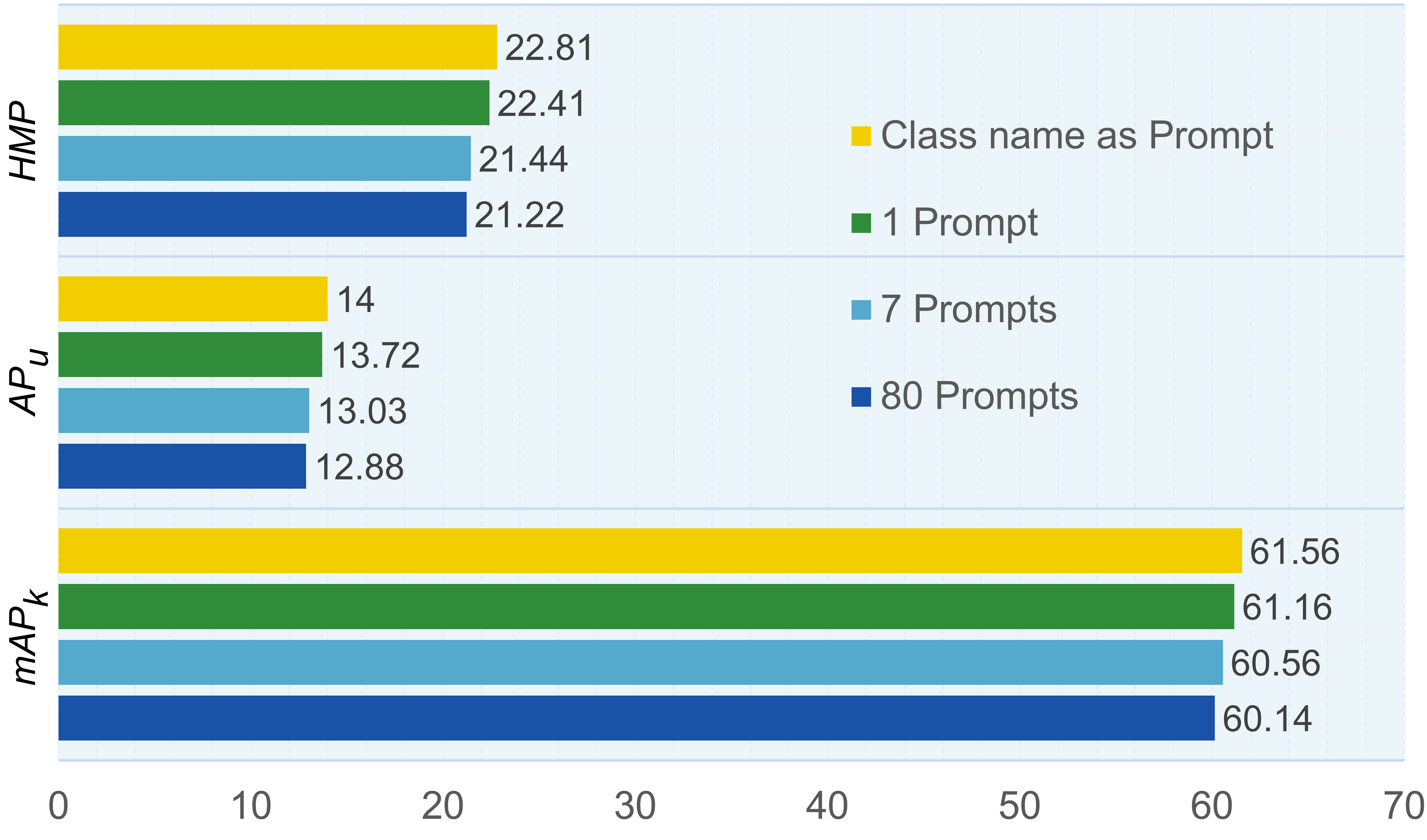}
    \caption{\small Effect of different prompts in CLIP-based text encoder of semantic clustering module on VOC-COCO-40 setting.}
    \label{fig:CLIP-Abl}
\end{figure}

In the proposed framework, we have introduced a semantic clustering module that utilizes a CLIP-based text encoder \cite{CLIP} to generate a 1024-dimensional text embedding. In contrast to the original CLIP approach \cite{CLIP} that uses a single prompt, seven prompts, or 80 prompts, we have utilized the class name as the prompt. To see the impact of using only the class name as a prompt, we conducted several ablation experiments in which the proposed framework is trained with different prompts in the semantic clustering module. The corresponding findings, depicted in Figure \ref{fig:CLIP-Abl}, indicate that the proposed framework with only class name as prompt performs better than other settings in terms of $mAP_k$, $AP_u$, and \textit{HMP} metrics.  
\begin{table*}[t!]
\caption{Comparison with SOTA methods on VOC-COCO-T1 setting on \textbf{Swin-T backbone}. The best-performing measures are highlighted with \textbf{bold font} while the second-best is highlighted with \underline{\textit{underlined italic font}}. * indicates the re-trained methods.} \label{tab:Sup_T1}
\begin{adjustbox}{width=.99\linewidth,center}
\begin{tabular}{|l|c|ccccc|ccccc|ccccc|}
\hline
\multicolumn{1}{|c|}{\multirow{2}{*}{Method}} & VOC    & \multicolumn{5}{c|}{VOC-COCO-20} & \multicolumn{5}{c|}{VOC-COCO-40}  & \multicolumn{5}{c|}{VOC-COCO-60}      \\ \cline{2-17} 
     & $mAP_k \uparrow$ &  $WI \downarrow$      &  $AOSE \downarrow$    &  $mAP_k \uparrow$   & $AP_u \uparrow$ & \textit{HMP} $\uparrow$ &  $WI \downarrow$      &  $AOSE \downarrow$    &  $mAP_k \uparrow$    & $AP_u \uparrow$ & \textit{HMP} $\uparrow$ &  $WI \downarrow$      &  $AOSE \downarrow$    &  $mAP_k \uparrow$    & $AP_u \uparrow$ & \textit{HMP} $\uparrow$ \\ \hline
Faster RCNN* \cite{FasterRCNN} & 78.74 & 11.39 & 21562 & 57.21 & 0.00 & 0.00 & 14.77 & 34074 & 53.73 & 0.00 & 0.00 & 12.13 & 39883 & 54.39 & 0.00 & 0.00 \\ \hline
DS* \cite{DS} & 78.08 & 9.58 & 16769 & 57.81 & 7.51 & 13.29 & 12.03 & 24946 & 54.54 & 5.35 & 9.74 & 9.69 & 27422 & 55.11 & 1.74 & 3.37 \\ \hline
PROSER* \cite{PROSER} & 78.94 & 13.96 & 19593 & 57.66 & \underline{\textit{16.38}} & \underline{\textit{25.51}} & 17.04 & 29567 & 54.29 & \underline{\textit{11.29}} & 18.69 & 13.62 & 33686 & 54.92 & \textbf{4.55} & \underline{\textit{8.40}} \\ \hline
OpenDet* \cite{OpenDet}    & \textbf{79.92}  & \underline{\textit{8.31}}  & \underline{\textit{12743}} & \textbf{59.79}  & 15.87 & 25.08 & \underline{\textit{10.40}}  & \underline{\textit{18925}}  & \textbf{57.19}   & 11.25 & \underline{\textit{18.80}} & \underline{\textit{9.12}}  & \underline{\textit{24073}} & \textbf{57.89} &  \underline{\textit{4.38}} & 8.14 \\ \hline
\rowcolor{LightCyan}\textbf{Our (proposed)}    & \underline{\textit{79.27}}  & \textbf{8.12}  & \textbf{10667} & \underline{\textit{58.87}}  & \textbf{16.93} & \textbf{26.30}  &   \textbf{9.90}   &   \textbf{15895}  &   \underline{\textit{56.24}}  &    \textbf{11.85} & \textbf{19.58} &  \textbf{8.73}  & \textbf{20924} & \underline{\textit{57.35}} & \textbf{4.55}  & \textbf{8.43} \\ \hline
\end{tabular}
\end{adjustbox}
\end{table*}
\begin{table*}
\caption{Comparison with SOTA methods on VOC-COCO-T2 setting on \textbf{Swin-T backbone}. The best-performing measures are highlighted with \textbf{bold font} while the second-best is highlighted with \underline{\textit{underlined italic font}}. * indicates the re-trained methods.}\label{tab:Sup_T2}
\begin{adjustbox}{width=.99\linewidth,center}
\begin{tabular}{|l|ccccc|ccccc|ccccc|}
\hline
\multirow{2}{*}{Methods} & \multicolumn{5}{c|}{VOC-COCO-n}   & \multicolumn{5}{c|}{VOC-COCO-2n}   & \multicolumn{5}{c|}{VOC-COCO-4n}  \\ \cline{2-16} 
& $WI \downarrow$      &  $AOSE \downarrow$    &  $mAP_k \uparrow$   & $AP_u \uparrow$ & \textit{HMP} $\uparrow$ & $WI \downarrow$      &  $AOSE \downarrow$    &  $mAP_k \uparrow$   & $AP_u \uparrow$ & \textit{HMP} $\uparrow$ & $WI \downarrow$      &  $AOSE \downarrow$    &  $mAP_k \uparrow$   & $AP_u \uparrow$ & \textit{HMP} $\uparrow$ \\ \hline
Faster RCNN* \cite{FasterRCNN} & 13.61 & 16941 & 70.31 & 0.00 & 0.00 & 21.04 & 33888 & 64.48 & 0.00 & 0.00 & 28.29 & 67729 & 57.41 & 0.00 & 0.00 \\ \hline
DS* \cite{DS} & 11.99 & 11404 & 69.48 & 5.59 & 10.35 & 18.98 & 22664 & 63.87 & 7.24 & 13.01 & 26.22 & 45162 & 56.79 & 8.56 & 14.88 \\ \hline
PROSER* \cite{PROSER} & 14.56 & 14742 & 69.85 & 11.63 & 19.94 & 22.96 & 29224 & 64.32 & 14.30 & 23.40 & 30.87 & 58593 & 56.85 & 16.31 & 25.35 \\ \hline
OpenDet* \cite{OpenDet}    & \textbf{9.21} & \underline{\textit{8896}} &  \textbf{75.28} & \underline{\textit{12.48}} & \underline{\textit{21.41}}  & \underline{\textit{15.46}} &  \underline{\textit{17665}} & \textbf{70.80}  & \underline{\textit{15.10}} & \underline{\textit{24.89}} & \underline{\textit{22.98}} & \underline{\textit{35365}} & \textbf{64.34}  & \underline{\textit{17.04}} & \underline{\textit{26.94}} \\ \hline
\rowcolor{LightCyan}\textbf{Our (proposed)}    & \underline{\textit{9.29}} & \textbf{7383} & \underline{\textit{74.04}}  & \textbf{13.36} & \textbf{22.64}  & \textbf{15.33} & \textbf{14706}  & \underline{\textit{69.72}} & \textbf{16.23} & \textbf{26.33} & \textbf{22.62} & \textbf{29600} & \underline{\textit{63.49}} & \textbf{17.97} & \textbf{28.01} \\ \hline
\end{tabular}
\end{adjustbox}
\end{table*}
\subsection{Effect of different thresholds in entropy thresholding evaluation mechanism}
\begin{figure}[t!]
    \centering
    \includegraphics[width = 0.99\linewidth, height=\linewidth]{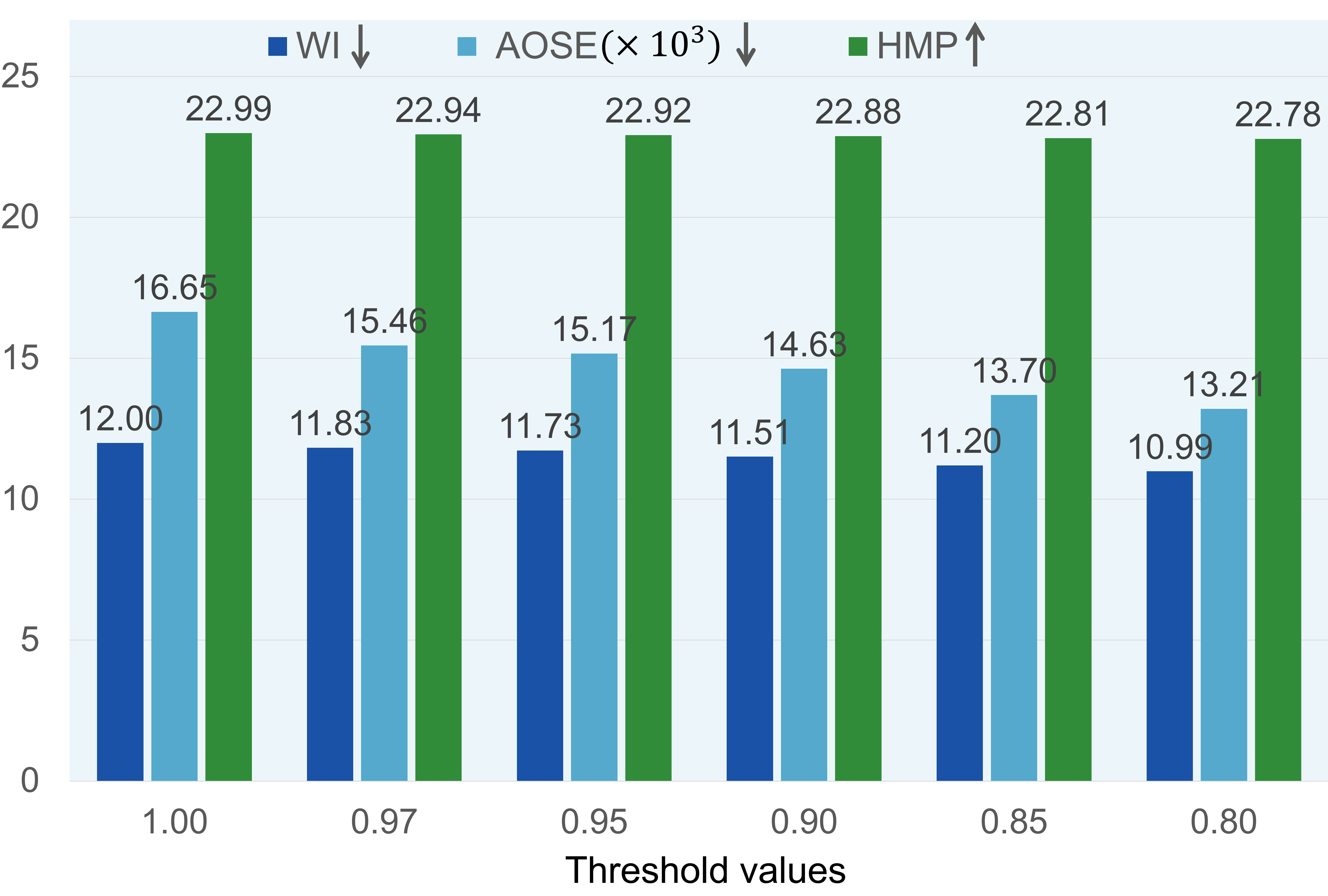}
    \caption{\small Effect of different thresholds in entropy thresholding mechanism on VOC-COCO-40 setting.}
    \label{fig:Entropy_Abl}
    \vspace{-1em}
\end{figure}

Figure \ref{fig:Entropy_Abl} depicts the impact of varying threshold values for entropy thresholding. Reducing the threshold value results in a decrease in the number of misclassified unknown instances, which leads to an improvement in $AOSE$. Simultaneously, the $WI$ is also improved by decreasing the threshold value. However, this reduction also coincides with a decrease in precision scores. The decline in $AP_u$ arises from the incompleteness of annotations in the COCO dataset, which results in numerous unknown predictions being classified as False Positives. As a result, there is a trade-off between achieving a favorable $AOSE$ score and maintaining a high precision score through entropy thresholding. We opt for a threshold of 0.85 for our experiments as it yields balanced performance across all metrics.

\section{ Experimental Results}\label{sup:result}
In addition to the experimental analysis presented in the main manuscript, we elaborate on some additional experimental results. 

\begin{table}[t!]
\caption{Comparison with SOTA methods on VOC-COCO-0.5n setting. This table is an extension of Table 2 in our main paper. The best-performing measures are highlighted in bold font, while the second-best is highlighted with an underlined italic font. * indicates the re-trained methods.}\label{tab:Sup_T2_0.5n}

\begin{adjustbox}{width=.99\linewidth,center}
\begin{tabular}{|l|ccccc|}
\hline
& $WI \downarrow$      &  $AOSE \downarrow$    &  $mAP_k \uparrow$   & $AP_u \uparrow$ & \textit{HMP} $\uparrow$  \\ \hline
\multicolumn{6}{|l|}{\textbf{ResNet50 as Backbone}} \\ \hline
Faster RCNN \cite{FasterRCNN}              & 9.25 & 6015 & \underline{\textit{77.97}}  & 0.00 & 0.00   \\ \hline
ORE \cite{ORE}                      & 8.39 & 4945 & 77.84  & 1.75 & 3.42\\ \hline
DS \cite{DS}                      & 8.30 & 4862 & 77.78  & 2.89  & 5.57\\ \hline
PROSER \cite{PROSER}                  & 9.32 & 5105 & 77.35  & 7.48  & 13.64\\ \hline
OpenDet \cite{OpenDet}                 & \underline{\textit{6.44}} & \underline{\textit{3944}} & \textbf{78.61}  & \underline{\textit{9.05}} & \underline{\textit{16.23}} \\ \hline
Openset RCNN \cite{OpenSetRCNN}             & 6.66 & 3993 & 77.85  & --- & ---  \\ \hline
\rowcolor{LightCyan} \textbf{Our (proposed)}          & \textbf{5.21} & \textbf{3363} & 76.06  & \textbf{12.70} & \textbf{21.77} \\ \hline
\multicolumn{6}{|l|}{\textbf{ConvNet-small as Backbone}} \\ \hline 
Faster RCNN* \cite{FasterRCNN}                   &8.64  & 5769 & \underline{\textit{82.68}}& 0.00 & 0.00  \\ \hline
DS* \cite{DS}                  & 8.23 & 5522 & 80.41  & 3.49 & 6.69 \\ \hline
PROSER* \cite{PROSER}     & 7.82 & 5054 &  82.00 & \underline{\textit{11.33}} & \underline{\textit{19.91}} \\ \hline
OpenDet* \cite{OpenDet}      &  \underline{\textit{5.30}}  &  \underline{\textit{3789}}  &  82.26   & 10.37 & 18.42  \\ \hline
\rowcolor{LightCyan} \textbf{Our (proposed)}          &  \textbf{5.05}  &  \textbf{3548}  &  \textbf{82.74}   & \textbf{13.96} & \textbf{23.89} \\ \hline
\multicolumn{6}{|l|}{\textbf{Swin-T as Backbone}} \\ \hline 
Faster RCNN* \cite{FasterRCNN}                   & 8.40  & 8471 & \underline{\textit{74.66}} & 0.00 & 0.00  \\ \hline
DS* \cite{DS}                  & 7.30 & 5753 & 73.81  & 4.03  & 7.64 \\ \hline
PROSER* \cite{PROSER}     & 9.19 & 7414 &  74.36 & \underline{\textit{9.61}} &  17.02\\ \hline
OpenDet* \cite{OpenDet}      &  \textbf{5.28}  &  \underline{\textit{4397}}  &  78.49   & 10.10 & \underline{\textit{17.90}}  \\ \hline
\rowcolor{LightCyan} \textbf{Our (proposed)}          &  \underline{\textit{5.46}}  &  \textbf{3717}  &  \textbf{77.57}   & \textbf{11.33} & \textbf{19.77} \\ \hline
\end{tabular}
\end{adjustbox}
\end{table}

\begin{figure*}[htb!]
    \centering
    \includegraphics[width = 0.97\linewidth, height=0.90\textheight]{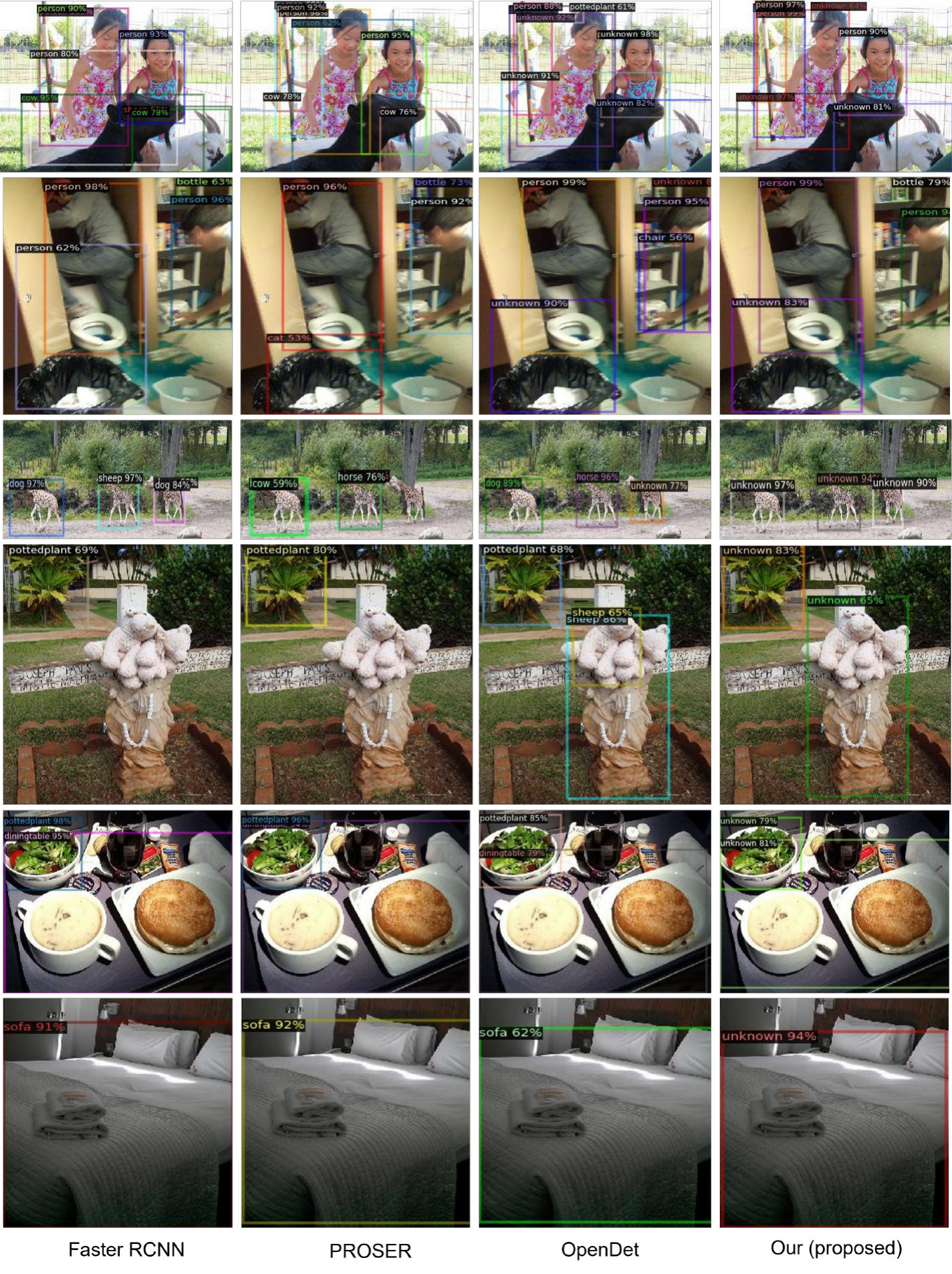}
    \caption{\small Visual comparison between our proposed and other methods. (Zoomed-in for better view)}
    \label{fig:visualize_supple}
\end{figure*}

\subsection{ Additional Results}
In addition to result comparison with existing OSOD works on ResNet50 and ConvNet backbone, we have also compared results on Swin-T \cite{SwinT} backbone. In Table \ref{tab:Sup_T1}, we present a comparison on the VOC-COCO-T1 setting. Here, we can see that the proposed method improves $WI$ by $2-5$\% and $AOSE$ by $2000-3000$ than previous best-performing PROSER \cite{PROSER} and OpenDet \cite{OpenDet} results in all dataset settings. We also show improvements as high as $5$\% on $AP_u$ and  $4$\% on \textit{HMP}, on VOC-COCO-40 compared to previous best-performing results. 
In VOC-COCO-60, the proposed method obtains a better $AP_u$ score of $4.55$ similar to PROSER \cite{PROSER}; however, it outperforms PROSER model in $mAP_k$ by a gain of $4.4$\%. 
Moreover, the comparison of VOC-COCO-T2 setting is presented in Table \ref{tab:Sup_T2}, where the proposed method performs better than other methods in terms of $AOSE$, $AP_u$ and \textit{HMP} metrics in all settings. We show a gain of $5-8$\% in $AP_u$, $4-6$\% in \textit{HMP} and improves $AOSE$ by $1500-5500$ than OpenDet \cite{OpenDet}. 


Due to limited space in our main paper, we also report the results on VOC-COCO-0.5n in Table \ref{tab:Sup_T2_0.5n} based on ResNet50, ConvNet and Swin-T backbones. Here, one can observe that the proposed method outperforms existing methods by a significant margin in all cases except in $mAP_k$ from the ResNet50 backbone-based comparison. 

\subsection{ Comparison on OWOD setting}
We evaluate the proposed method in the context of OWOD setting, i.e., task 1 as suggested in \cite{ORE} and compare the results against existing methods, presented in Table \ref{tab:Sup_ORE}. This analysis reveals that the proposed method performs better when employed with a ResNet50 backbone than other methods in terms of $WI$ and $AOSE$. Furthermore, when used with a ConvNet backbone, our proposed method improves the performance further and obtains significant performance than other methods.

\begin{table}[h!]
\caption{Comparison with OWOD based task 1 evaluation setting \cite{ORE}. The best-performing measures are highlighted with \textbf{bold font}. $\dagger$ indicates results obtains from OpenDet \cite{OpenDet} paper, while $\dagger\dagger$ indicates results from Openset-RCNN \cite{OpenSetRCNN} paper.} \label{tab:Sup_ORE}

\begin{adjustbox}{width=.9\linewidth,center}
\begin{tabular}{|l|ccc|}
\hline
\multirow{2}{*}{Method} & \multicolumn{3}{c|}{OWOD-Task-1} \\
\cline{2-4}
& $WI \downarrow$      &  $AOSE \downarrow$    &  $mAP_k \uparrow$   \\ \hline
Faster R-CNN (Oracle)$^\dagger$ \cite{FasterRCNN} &  4.27     &   6862    &   60.43   \\ \hline
Faster R-CNN$^\dagger$ \cite{FasterRCNN}  &   6.03    &   8468    &  58.81    \\ \hline
ORE$^\dagger$ \cite{ORE}  &   5.11    &   6833     &    56.34    \\ \hline
OW-DETR$^{\dagger\dagger}$ \cite{OW-DETR} &   5.71    &  10240      &  59.21     \\ \hline
PROB \cite{PROB}  &  ---     &   ---     &    59.50   \\ \hline
OpenDet \cite{OpenDet}    &  4.44     & 5781       &    59.01   \\ \hline
Openset-RCNN \cite{OpenSetRCNN}  &  4.67     &  5403      &  59.34   \\ \hline
\rowcolor{LightCyan} \textbf{Our (ResNet50)}  & 3.76 & 5145 &  57.44   \\ \hline 

\rowcolor{LightCyan} \textbf{Our (ConvNet)}  & \textbf{3.52}  &  \textbf{4616} &  \textbf{61.51}    \\ \hline
\end{tabular}
\end{adjustbox}
\end{table}

\subsection{ Additional Qualitative Results}
In addition to quantitative analysis, we have provided qualitative results in Figure~\ref{fig:visualize_supple} to demonstrate the improvement of our method over baseline methods such as Faster RCNN \cite{FasterRCNN}), PROSER \cite{PROSER} and previous best-performing OpenDet \cite{OpenDet}. It can be visualized that the proposed method accurately classifies unknown objects that are semantically closer to known classes, which other methods fail to do. For example, Faster R-CNN \cite{FasterRCNN} and PROSER \cite{PROSER} misclassify `goat' as `cow' due to their semantic similarity (see $1^{st}$ row of Figure~\ref{fig:visualize_supple}). However, our model, having learned semantic-based clusters, correctly labels `goat' as the `unknown' class. It can also be observed that other methods misclassify the `giraffe' as either `cow', `sheep', `dog' or 'horse' as depicted in $3^{rd}$ row in Figure~\ref{fig:visualize_supple}. In contrast, our proposed method accurately identifies it as `unknown'. Similarly, other models misclassify `bed' as ‘sofa’ due to their semantic similarity. At the same time, the proposed method predicts it accurately as an unknown class (as illustrated in the last row in Figure~\ref{fig:visualize_supple}).

\subsection{ Failure Case Analysis}
\begin{figure}[h!]
    \centering
    \includegraphics[width = 0.99\linewidth]{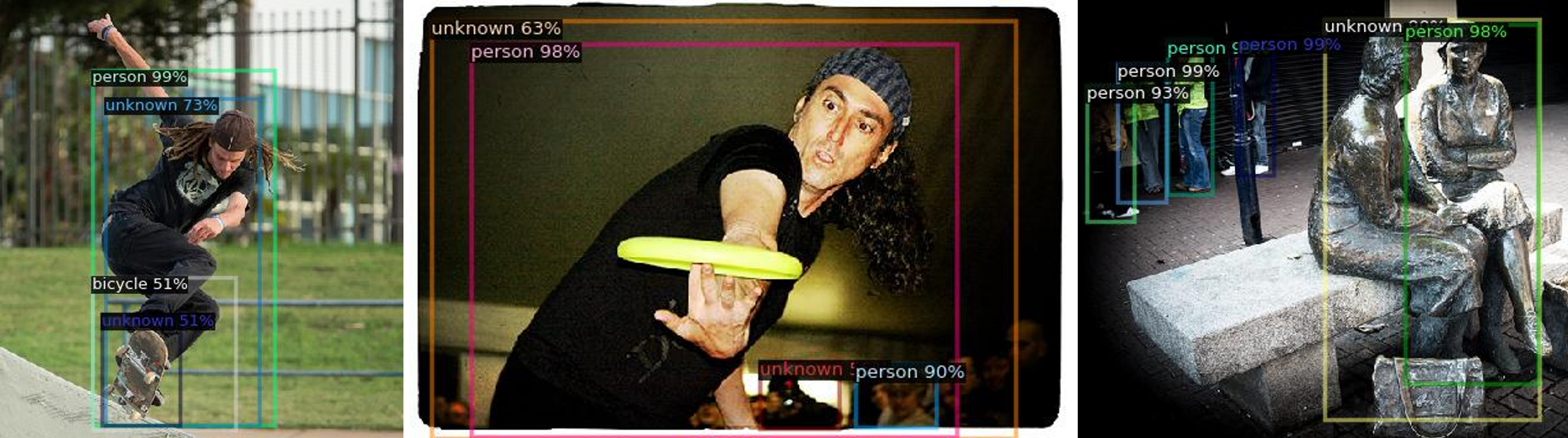}
    \caption{\small failure cases.}
    \label{fig:Failure}
    \vspace{-1em}
\end{figure}

In Figure \ref{fig:Failure}, we present several instances where our model fails to perform well. The proposed framework detects false positive 'unknown' objects in all three images. We posit that this problem may arise due to a limitation of the object focus loss and its tendency to promote additional unknown detection. As a result, in certain cases, this mechanism may detect objects that are not even present.

\end{document}